\documentclass[10pt]{article} 
\usepackage[accepted]{tmlr}

\usepackage{amsmath,amsfonts,bm}

\def\figref#1{Fig.~\ref{#1}}

\def\eqref#1{equation~\ref{#1}}

\def\1{\bm{1}}

\DeclareMathAlphabet{\mathsfit}{\encodingdefault}{\sfdefault}{m}{sl}
\SetMathAlphabet{\mathsfit}{bold}{\encodingdefault}{\sfdefault}{bx}{n}

\usepackage{hyperref}
\usepackage{url}

\usepackage[utf8]{inputenc} 
\usepackage[T1]{fontenc}    
\usepackage{booktabs}       
\usepackage{amsfonts}       
\usepackage{nicefrac}       
\usepackage{microtype}      
\usepackage{xcolor}         
\usepackage{bm}

\usepackage{graphicx}
\usepackage{amsmath,amssymb}
\usepackage{bbding}
\usepackage{tabularx}
\usepackage{booktabs}
\usepackage{multirow}
\usepackage{xspace}
\usepackage[acronym]{glossaries}
\usepackage[capitalize]{cleveref}
\usepackage{float}
\usepackage{wrapfig}

\newcommand{\todo}[1]{\textcolor{black}{#1}}
\newcommand{\rebuttal}[1]{\textcolor{black}{#1}}
\newcommand{\colrebuttal}[1]{\color{black}{#1}}
\newcommand{\rebuttaltmlr}[1]{\textcolor{black}{#1}}

\title{3D-Aware Video Generation}

\author{Sherwin Bahmani$^{1}$
  \space\space
  Jeong Joon Park$^{2}$
  \space\space
  Despoina Paschalidou$^{2}$
  \space\space
  Hao Tang$^{1}$\\
  \space\space
  Gordon Wetzstein$^{2}$
  \space\space
  Leonidas Guibas$^{2}$
  \space\space
  Luc Van Gool$^{1,3}$
  \space\space
  Radu Timofte$^{1,4}$\\
  \textnormal{$^{1}$ETH Zürich \space\space$^{2}$Stanford University\space\space $^{3}$KU Leuven\space\space $^{4}$University of W\"urzburg}
}

\def\month{06}  
\def\year{2023} 
\def\openreview{\url{https://openreview.net/forum?id=SwlfyDq6B3}} 

\begin{document}

\maketitle

\begin{abstract}


Generative models have emerged as an essential building block for many image synthesis and  editing tasks. Recent advances in this field have also enabled high-quality 3D or video content to be generated that exhibits either multi-view or temporal consistency. 
With our work, we explore 4D generative adversarial networks (GANs) that learn unconditional generation of 3D-aware videos. 
By combining neural implicit representations with time-aware discriminator, we develop a GAN framework that synthesizes 3D video supervised only with monocular videos. We show that our method learns a rich embedding of decomposable 3D structures and motions that enables new visual effects of spatio-temporal renderings while producing imagery with quality comparable to that of existing 3D or video GANs. 

\end{abstract}

\begin{figure}[h!]
    \centering
    \includegraphics[width=\textwidth]{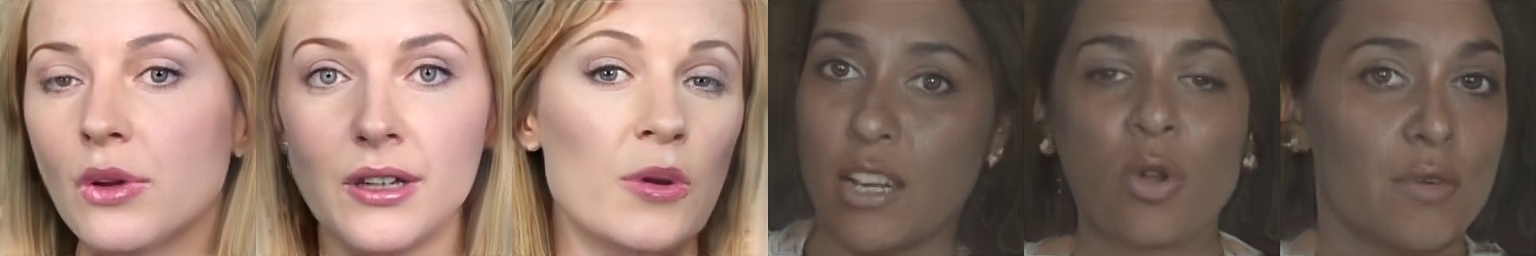}
        \vspace{-0.7em}
    \caption{{\bf{3D-Aware video generation.}} 
    We show multiple frames and viewpoints of two 3D videos, generated using our model trained on the FaceForensics dataset~\citep{Rossler2019ICCV}. Our 4D GAN generates 3D content of high quality while permitting control of time and camera extrinsics. Video results can be viewed on our website: \url{https://sherwinbahmani.github.io/3dvidgen}
    }
    \label{fig:teaser}
    \vspace{-0.5em}
\end{figure}

\section{Introduction}
\label{sec:intro}

Recent advances in generative adversarial networks
(GANs)~\citep{Goodfellow2014NIPS} have led to artificial synthesis of photorealistic images~\citep{Karras2019CVPR,Karras2020CVPR,Karras2021NIPS}. These methods have been extended to enable unconditional generation of high-quality videos~\citep{Chen2021NIPS, Yu2022ICLR} and multi-view-consistent 3D scenes~\citep{Gu2022ICLR,Chan2022CVPR,Or2022CVPR}. However, despite important applications in visual effects, computer vision, and other fields, no generative model has been demonstrated that is successful in synthesizing 3D videos to date.

We propose the first 4D GAN that learns to generate multi-view-consistent video data from single-view videos. For this purpose, we develop a 3D-aware video generator to synthesize 3D content that is animated with learned motion priors, and permits viewpoint manipulations. Two key elements of our framework are a time-conditioned 4D generator that leverages emerging neural implicit scene representations~\citep{Park2019CVPR,Mescheder2019CVPR,Mildenhall2020ECCV} and a time-aware video discriminator. Our generator takes as input two latent code vectors for 3D identity and motion, respectively, and it outputs a 4D neural fields that can be queried continuously at any spatio-temporal $xyzt$ coordinate. The generated 4D fields can be used to render realistic video frames from arbitrary camera viewpoints. To train the 4D GAN, we use a discriminator that takes two randomly sampled video frames from the generator (or from real videos) along with their time differences to score the realism of the motions. Our model is trained with an adversarial loss where the generator is encouraged, by the discriminator, to render realistic videos across all sampled camera viewpoints. 

\rebuttal{Our contributions are following: i) We introduce the first 4D GAN which generates 3D-aware videos supervised only from single-view 2D videos.
ii) We develop a framework combining implicit fields with a time-aware discriminator, that can be continuously rendered for any $xyzt$ coordinate.
iii) We evaluate the effectiveness of our approach on \rebuttaltmlr{various} video datasets. We show that the trained 4D GAN is able to synthesize plausible videos that allows viewpoint changes, whose visual and motion qualities are competitive against the state-of-the-art 2D video GANs' outputs.
}

\vspace{-0.2em}
\section{Related Work}
\vspace{-0.1em}
\label{sec:related}

In this section, we discuss the most relevant literature on image and video synthesis as
well as neural implicit representations in the context of 2D and 3D content generation. See appendix for a more complete list of references.

\paragraph{GAN-based Image Synthesis}%
Generative Adversarial Networks (GANs) \citep{Goodfellow2014NIPS} have demonstrated impressive results on multiple synthesis tasks such as image generation \citep{Brock2019ICLR, Karras2019CVPR,Karras2020CVPR}, image editing \citep{Wang2018CVPRa, Shen2020CVPR, Ling2021NIPS} and image-to-image translation
\citep{Isola2017CVPR, Zhu2017ICCV, Choi2018CVPR}. To allow for increased
controllability, during the image synthesis process, several recent works have
proposed to disentangle the underlying factors of variation \citep{Reed2014ICML,Chen2016NIPS, Lee2020ECCV,Shoshan2021ICCV} or rely on pre-defined templates \citep{Tewari2020TOG,Tewari2020CVPR}.  However, since most of these methods operate on 2D images, they often lack physically-sound control in terms of viewpoint manipulation. In this work, we advocate modelling both the image and
the video generation process in 3D in order to ensure controllable generations.

\paragraph{Neural Implicit Representations}%
Neural Implicit Representation (NIR) \citep{Mescheder2019CVPR, Park2019CVPR, Chen2019CVPRb} have been extensively employed in various generation tasks due to their continuous, efficient, and differentiable nature. These tasks include 
3D reconstruction of objects and scenes, novel-view synthesis of static and dynamic scenes, inverse graphics, and video representations \citep{Jiang2020CVPR, Chibane2020CVPR, Sitzmann2020NIPS, Barron2021ICCV, Sajjadi2022CVPR, Park2021ICCV, Li2021CVPR, Niemeyer2020CVPR, Chen2021NIPS}. Among the most widely used NIRs are Neural Radiance Fields (NeRFs)
\citep{Mildenhall2020ECCV} that combine 
NIRs with volumetric rendering to enforce 3D consistency while performing
novel view synthesis.
In this work, we employ a generative variant of
NeRF \citep{Gu2022ICLR} and combine it
with a time-aware discriminator to learn a generative model of videos from
unstructured videos. 

Closely related to our work are recent approaches that try to control the motion and the pose of scenes  \citep{Lin2022ARXIV,Liu2021TOG,Zhang2021TOG,Ren2021ICCV}. In particular, they focus on transferring or controlling the motion of their target objects instead of automatically generating plausible motions. They often use networks overfitted to a single reconstructed scene \citep{Chen2021ARXIV,Zhang2021TOG} or rely on pre-defined templates of human faces or bodies \citep{Liu2021TOG,Ren2021ICCV}. In our work, we build a 4D generative model that can automatically generate diverse 3D content along with its plausible motion without using any pre-defined templates.

\paragraph{3D-Aware Image Generations}%
Another line of research investigates how 3D representations can be incorporated in generative settings for improving the image quality
\citep{Park2017CVPR, Nguyen-Phuoc2018NIPS} and increasing the control
over various aspects of the image formation process \citep{Gadelha2017THREEDV, Chan2021CVPR, Henderson2019IJCV}. 
Towards this goal, several works \citep{Henzler2019ICCV,Nguyen-Phuoc2019ICCV,Nguyen-Phuoc2020ARXIV} proposed to train 3D-aware GANs from a set of unstructured images using voxel-based representations. However, due to the low voxel resolution and the inconsistent view-controls stemming from the use of pseudo-3D structures that rely on non-physically-based 2D-3D conversions, these methods tend to generate images with artifacts and struggle to generalize in real-world scenarios.
More recent approaches rely on volume rendering to generate 3D objects \citep{Schwarz2020NIPS,Chan2021CVPR,Niemeyer2021CVPR}. 
Similarly, \citet{Zhou2021ARXIV}, \citet{Chan2021CVPR}, and
\citet{DeVries2021ICCV} explored the idea of combining NeRF with GANs for designing
3D-aware image generators. 
Likewise, StyleSDF \hbox{\citep{Or2022CVPR}} and StyleNeRF
\citep{Gu2022ICLR} proposed to combine an MLP-based volume renderer with a style-based generator \citep{Karras2020CVPR} to
produce high-resolution 3D-aware images. \cite{Deng2022CVPR} explored learning a
generative radiance field on 2D manifolds and \cite{Chan2022CVPR} introduced a
3D-aware architecture that exploits both implicit and explicit representations. To improve view-consistency, EpiGRAF \citep{skorokhodov2022epigraf} proposes patch-based training to discard the 2D upsampling network, while VoxGRAF \citep{schwarz2022voxgraf} uses sparse voxel grids for efficient rendering without a superresolution module. \rebuttaltmlr{AniFaceGAN \citep{wu2022anifacegan} proposes an animatable 3D-aware GAN for face animation generation supervised by image collections.}.
In contrast to this line of research that focuses primarily on 3D-aware
image generation, we are interested in \emph{3D-aware video generation}. In
particular, we build on top of StyleNeRF \citep{Gu2022ICLR} to allow control on the 3D camera pose during the video synthesis. To the best
of our knowledge, this is the first work towards 3D-aware video generation trained from unstructured 2D data.

\paragraph{GAN-based Video Synthesis}%
Inspired by the success of GANs and adversarial training on
photorealistic image generation, researchers  shifted their attention
to various video synthesis tasks \citep{Tulyakov2018CVPR, Holynski2021CVPR, Tian2021ICLR}. Several works pose the video synthesis as an autoregressive video prediction
task and seek to generate future frames conditioned on the previous using either
recurrent~\citep{Kalchbrenner2017ICML, Walker2021ARXIV} or
attention-based~\citep{Rakhimov2020ARXIV,
Weissenborn2020ICLR, Yan2021ARXIV} models. 
Other approaches
\citep{Saito2017ICCV, Tulyakov2018CVPR, Aich20202CVPR} tried to
disentangle the motion from the image
generation during the video synthesis process. 
To facilitate
generating high-quality frames, \citet{Tian2021ICLR} and \citet{Fox2021ARXIV} employed a
pre-trained image generator of \citet{Karras2020CVPR}. Recently, LongVideoGAN \citep{brooks2022generating} has investigated synthesizing longer videos of more complex datasets.
Closely related to our method are the recent work of DIGAN~\citep{Yu2022ICLR} and StlyeGAN-V~\citep{Skorokhodov2022CVPR} that generate videos at continuous time steps, without conditioning on previous frames.
DIGAN~\citep{Yu2022ICLR} employs an NIR-based image generator
\citep{Skorokhodov2021CVPR} for learning continuous videos and introduces two discriminators:
the first discriminates the realism of each frame
and the second operates on image pairs and seeks to determine the realism of
the motion. Similarly, StyleGAN-V~\citep{Skorokhodov2022CVPR} employs a style-based GAN
\citep{Karras2020CVPR} and a single discriminator that operates on sparsely sampled frames. In contrast to \citet{Yu2022ICLR} and \citet{Skorokhodov2022CVPR}, we focus on \emph{3D-aware video generation}. In particular, we build on top of StyleNeRF~\citep{Gu2022ICLR} and DIGAN~\citep{Yu2022ICLR} and demonstrate the ability of our model to render high quality videos from diverse viewpoint angles. Note that this task is not possible for prior works that do not explicitly model the image formation process in 3D.

\vspace{-0.2em}
\section{Method}
\vspace{-0.1em}
\label{sec:method}

\begin{figure*}[t]
    \centering
    \includegraphics[width=\textwidth]{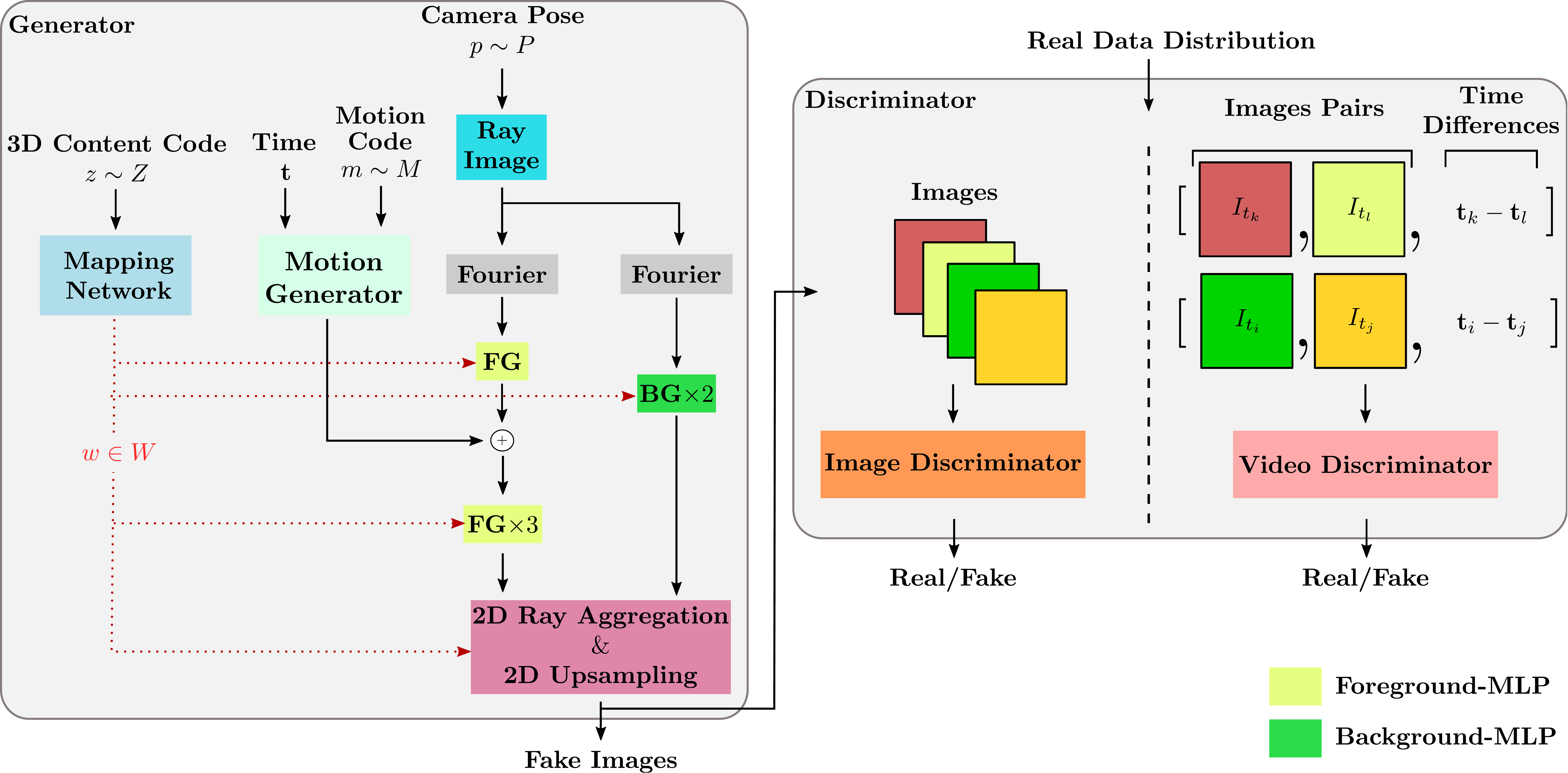}
    \vspace{-0.5em}
\caption{{\bf{Model architecture.}} The generator (left) takes 3D content and motion codes and the query camera view to render an RGB image. Given a camera pose, we construct a ray image, from which we sample $xyz$ positions along the rays to be passed to the fourier feature layer. The content code is transformed into $w$ and modulate the intermediate features of MLP layers. The motion code, along with the time query $t$, is processed with the motion generator and added to the generator branch. The video discriminator takes two random frames from a video and their time difference and outputs real/fake prediction. The image discriminator takes individual frames and outputs real/fake label. }
    \label{fig:model_architecture}
    \vspace{-1em}
\end{figure*}

The two main components of our 4D GAN framework are a time-conditioned neural scene generator and a time-aware discriminator. 
Our generator networks take as input two independent noise vectors, $\bm{m}$ and $\bm{z}$, that respectively modulate the motion and the content of the 4D fields. 
To render an image at a specific time step $t$, we sample the camera extrinsics according to dataset-dependent distribution and conduct volume rendering through the time-conditioned radiance and density fields. 
Our time-aware discriminator measures the realism of a pair of frames, given their time difference, to promote plausible 3D video generation. The overview of our pipeline can be found in \figref{fig:model_architecture}.

\subsection{Time-Conditioned Implicit Fields}
We build on top of existing coordinate-based neural generators \citep{Gu2022ICLR,Or2022CVPR} to model continuous implicit fields using a Multi-layer Perceptron (MLP) that outputs a density $\sigma(\bm{x},t;\bm{z},\bm{m})$ and appearance feature $\bm{f}(\bm{x},t,\bm{d};\bm{z},\bm{m})$ for a given spatio-temporal query $(\bm{x},t),$ and view direction $\bm{d}.$ Here, $\bm{z}$ and $\bm{m}$ are 3D content and motion latent vectors, respectively. 



We process a motion latent vector $\bm{m}$ sampled from a unit sphere with an MLP conditioned on time step $t\in [0,1],$ to obtain a final motion vector \rebuttal{$\bm{n}(\bm{m},t)$} via multiplicative conditioning:
\begin{equation}
    \rebuttaltmlr{\bm{n}(\bm{m},t)= \psi^3 \circ \psi^2 \circ (t \cdot (\psi^1 (\bm{m}))),}
\end{equation}
where $\psi^i$ is a fully connected layer. Leaky ReLU activation is applied between the layers. Note that our use of continuous time variable $t$ allows rendering the video at an arbitrary frame rate, unlike autoregressive models that can only sample discrete time steps.

The resulting motion vector $\bm{n}(\bm{m},t)$ is then used to control the output of the generator MLP  $g_{\bm{w}}(\bm{x},\bm{n})$ along with the modulation parameters $\bm{w}(\bm{z})$ computed from the 3D content vector $\bm{z}$:
\begin{equation} \label{eq: generator_mlp}
    g_{\bm{w}}(\bm{x},\bm{n})=\phi_{\bm{w}}^k \circ \phi_{\bm{w}}^{k-1} \circ \ldots \phi_{\bm{w}}^2 \circ (\bm{n}(\bm{m},t)+\phi_{\bm{w}}^1 \circ \gamma(\bm{x})),
\end{equation}
where $\phi^i_{\bm{w}}$ is a fully connected layer (with Leaky ReLU activations between the layers) whose weights are modulated by the style vector $\bm{w}$, following the style-based generation techniques of \citet{Gu2022ICLR} and \citet{Karras2020CVPR}. 
The style vector is produced with a mapping MLP network \rebuttal{$\zeta$: $\bm{w}=\zeta(\bm{z})$}, where $\bm{z}$ is sampled from the surface of unit sphere. $\gamma(\bm{x})$ is  a positional encoding vector of a spatial query $\bm{x}$ that induces sharper appearances, following~\cite{Mildenhall2020ECCV}. We omit positional encoding on the time step $t$, as the empirical results did not improve (see Sec.~\ref{sec:exp_ablations}). Note that the additive conditioning allows us to easily disable  the influence of  motion by setting $\bm{n}$ to be 0, which promotes training on both image and video data when necessary.

The density value at $\bm{x}$ is then computed by passing the feature $g_{\bm{w}}(\bm{x},\bm{n})$ to a two layer MLP \rebuttal{$\varphi_\sigma$}:
\begin{equation}
\rebuttaltmlr{\sigma(\bm{x},t;\bm{z},\bm{m})=\rebuttaltmlr{\varphi_\sigma} (g_{\bm{w}(\bm{z})}(\bm{x},\bm{n}(t,\bm{m})))}
\end{equation}


\paragraph{Image Rendering}
\vspace{-0.2em}
To render an image from a virtual camera with pose $\bm{p}$, we compute an appearance feature for a ray $\bm{r}(l)=\bm{o}+l\bm{d}$ going through each pixel that emanates from the camera focus $\bm{o}$ towards the direction $\bm{d}.$
Specifically, we approximate the volume rendering process with discrete point sampling along the rays and process the aggregated ray features with an MLP, denoted $h_{\bm{w}}$, conditioned on the view direction $\bm{d}:$
\begin{align}\label{eq:volume_rendering}
    \rebuttaltmlr{\bm{f}(\bm{r},t, \bm{d};\bm{z},\bm{m})=h_{\bm{w}} \circ \int_{l_i}^{l_f} T(l) \sigma(\bm{r}(l),t)g_{\bm{w}}(\bm{r}(l),\gamma(\bm{d}),\bm{n})dl,}
\end{align}
\vspace{-0.5em}

$\text{where } T(l)=\exp(-\int_{l_i}^l \sigma(\bm{r}(s),t)ds ).$ The volume rendering of Eq.~\ref{eq:volume_rendering} involves millions of MLP queries and thus becomes quickly intractable with high resolution images. To reduce the computational overhead, we adopt the 2D upsampling CNN of StyleNeRF~\citep{Gu2022ICLR} to convert the low-resolution volume rendering results of $\bm{f}$ into a high-resolution RGB image $\mathcal{I}$:
\begin{equation}\label{eq: upsample}
    \mathcal{I}_{\bm{p}}(t;\bm{z},\bm{m})=\text{CNN}_{\text{up}}(\bm{f}(\mathcal{R}_{\bm{p}},t, \bm{d};\bm{z},\bm{m})),
\end{equation}
where $\mathcal{R}_{\bm{p}}$ denotes an image composed of rays from camera pose $\bm{p}$, with slight abuse of notation.

\paragraph{Background and Foreground Networks}%
\vspace{-0.2em}
Videos consist of static background and moving foreground, hence it would be ineffective to model both foreground and background using one network and motion code. We follow the inverse sphere parameterization of \cite{Zhang2020ARXIV} to model the background with a second MLP network of $g_w$ that is modulated only with the content vector.


\subsection{Training}
\label{sec:method_training}
\vspace{-0.2em}
We train our 4D generator via adversarial loss, leveraging time-aware discriminators from the video GAN literature. The key idea of our training is to encourage the generator to render realistic video frames for all sampled viewpoints and time steps, by scoring their realism with the discriminators. 

\paragraph{Time-Aware Discriminator}
\vspace{-0.2em}
Unlike autoregressive generators, our continuous 4D generator can render frames at an arbitrary time step without knowing the 'past' frames. 
This feature allows using efficient time-aware discriminators \citep{Skorokhodov2022CVPR,Yu2022ICLR} that only look at sparsely sampled frames as opposed to the entire video that would require expensive 3D convolutions to process.
We adopt the  2D CNN discriminator $D_{\text{time}}$ of DIGAN \citep{Yu2022ICLR} to score the realism of the generated motion from two sampled frames. Specifically, we render a pair of frames $\mathcal{I}_{\bm{p}}(t_1;\bm{z},\bm{m})$ and $\mathcal{I}_{\bm{p}}(t_2;\bm{z},\bm{m})$ from the same 3D scene and camera pose.
 The input to the discriminator $D_{\text{time}}$ is a concatenation of the two RGB frames along with the time difference between the frames expanded to the image resolution $\mathcal{I}_{\text{repeat}}(t_2-t_1)$: 
\begin{equation}
   D_{\text{time}}: \left[ \mathcal{I}_{\bm{p}}(t_1;\bm{z},\bm{m}), \mathcal{I}_{\bm{p}}(t_2;\bm{z},\bm{m}), \mathcal{I}_{\text{repeat}}(t_2-t_1) \right] \to \mathbb{R}, \quad \text{where } t_2 > t_1.
\end{equation}
For real videos, we similarly pass a random pair of  frames along with their time difference to  $D_{\text{time}}$.

\paragraph{Single Image Discriminator}\vspace{-0.2em}
 In theory, the time-aware discriminator $D_{\text{time}}$ should be able to simultaneously measure the realism of the motion of the input pair and that of the individual frames. However, we empirically observe that training a separate discriminator that specializes on single frame discrimination improves quality (see Sec.~\ref{sec:exp_ablations}). We therefore adopt another discriminator $D_{\text{image}}$ that scores realism of individual images: $D_{\text{time}}: \mathcal{I}_{\bm{p}}(t;\bm{z},\bm{m})\to \mathbb{R}.$ Following 3D GAN approaches \citep{Gu2022ICLR,Or2022CVPR, Chan2022CVPR}, we use StyleGAN2 \citep{Karras2020CVPR} discriminator architecture without modifications.

\paragraph{Loss Functions}
Our training objectives include the adversarial losses from the two discriminators along with the R1 \citep{Mescheder2018ICML} and NeRF-Path regularizations \citep{Gu2022ICLR}:
\begin{equation}\label{eq: objective}
    \mathcal{L}(D, G) = \mathcal{L}_{\text{adv}}(D_{\text{time}}, G)+\mathcal{L}_{\text{adv}}(D_{\text{image}}, G)+\lambda_1 \mathcal{L}_{\text {R1 }}(D_{\text{time}},D_{\text{image}})+\lambda_2 \mathcal{L}_{\text {NeRF-path }}(G),
\end{equation}
where $G$ denotes the entire image generator machinery in Eq.~\ref{eq: upsample}, and $\lambda$'s are balancing parameters.
To compute the adversarial losses we use the non-saturating objective. 
Note that our networks are trained end-to-end without the progressive growing strategy. More details are provided in the supplementary.



\begin{figure*}[ht!]
    \centering
    \includegraphics[width=\textwidth]{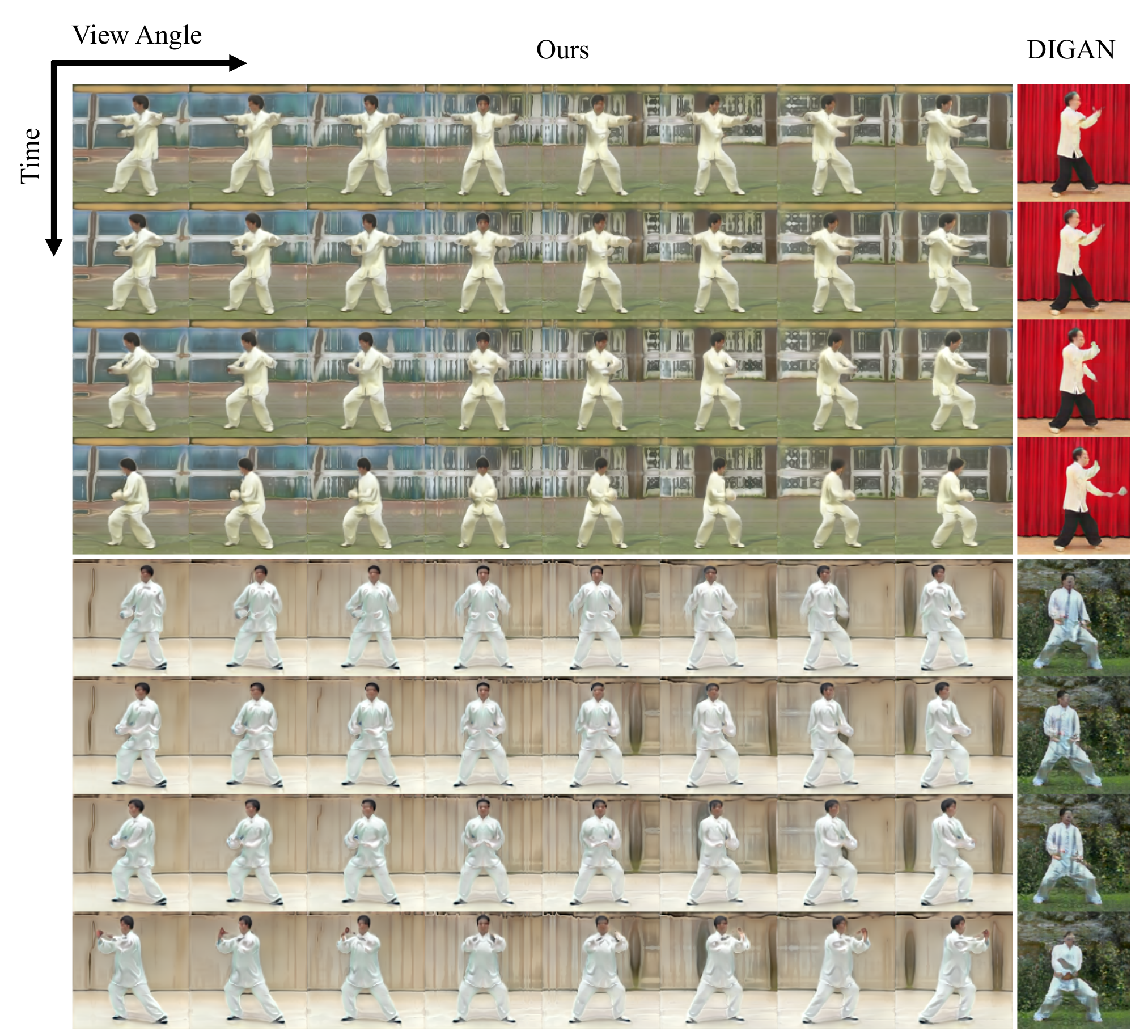}
    \vspace{-1.5em}
    \caption{
    {\bf{Qualitative results on TaiChi Dataset.}} We visualize the spatio-temporal renderings of two scenes sampled from our 4D GAN, where the horizontal axis indicates change of view angles while the vertical axis indicates progress of time. The rightmost column shows two videos sampled from DIGAN \citep{Yu2022ICLR} that can only be viewed from a fixed camera angle. Our method extends DIGAN in the spatial dimension while producing frames of comparable quality.
    }\label{fig: taichi1}
    \vspace{-1.0em}
\end{figure*}

\paragraph{Training on Image and Video Datasets}
The 3D structure of our generator emerges without 3D supervision by enforcing adversarial loss from sampled camera views. Thus, it is crucial that our video dataset features a diverse set of view angles. However, we notice that many of the popular video GAN datasets feature narrow range of viewpoints (e.g., FaceForensics). To address this issue, we seek to leverage existing 2D image datasets that are typically greater in quantity and diversity. 

We explore two options: (1) We pre-train our generator model $G$ on an image dataset and fine-tune it on a video dataset. During pre-training, we ignore the temporal components and sample the 3D content only. Such training can be done seamlessly by simply setting the motion vector $\bm{n}(\bm{m},t)$ of Eq.~\ref{eq: generator_mlp} to be zero. After pre-training, we unfreeze the temporal components and minimize the whole objective (Eq.~\ref{eq: objective}) on a video dataset. (2) We train our generator on image and video datasets simultaneously. 
\rebuttal{We refer the reader to Sec. \ref{sec:image_video_datasets} in our supplementary for more details.}  

\label{sec:method_implementation}

\section{Experiments}
\label{sec:experiments}
\vspace{-0.5em}

We conduct experiments to demonstrate the effectiveness of our approach in generating 3D-aware videos, focusing on the new visual effects it enables and the quality of generated imagery. Moreover, we conduct extensive ablation studies on the design components of our model. 
\vspace{-0.5em}

\subsection{Experimental Setup}\label{Sec: experiment setups}

\vspace{-0.2em}
\paragraph{Datasets}
We evaluate our approach on three publicly available, unstructured video datasets: the FaceForensics \citep{Rossler2019ICCV}, the MEAD \citep{Wang2020ECCV}, and the TaiChi \citep{Siarohin2019NIPS} dataset. 
FaceForensics contains 704 training videos of human faces sourced from YouTube. 
While this dataset containing in-the-wild videos makes it a great testbed for synthesis tasks, many of its videos are captured from frontal views with limited view diversity.
On the other hand, the MEAD dataset contains shorter videos capturing faces from discrete 7  angles, from which we randomly subsample 10,000. Note that we ignore the identity correspondences of the videos and treat them as independent unstructured videos. The TaiChi dataset contains 2942 in-the-wild videos of highly diverse TaiChi performances sourced from the internet. Following DIGAN \citep{Yu2022ICLR}, we use every fourth frame to make the motion more dynamic. Finally, we provide additional experiments on the SkyTimelapse dataset \citep{xiong2018learning} in the supplementary. 
\vspace{-1em}

\paragraph{Metrics}
Following existing video GAN methods, we use  Frechet
Video Distance (FVD) \citep{Unterthiner2018ARXICV} as our main metric for measuring realism of the generated  motion sequences.
In particular, we use the FVD protocol of \citet{Skorokhodov2022CVPR} that alleviates the inconsistency issues of the original FVD implementation \citep{Unterthiner2018ARXICV}.
Moreover,  we consider the following additional metrics: Average Content Distance (ACD) \citep{Tulyakov2018CVPR}, which measures temporal consistency of a video and CPBD \citep{narvekar2011no}, which measures the sharpness of an image. We also measure the face identity consistency (ID) across viewpoints by computing ArcFace \citep{deng2019arcface} cosine similarity, following \cite{Chan2022CVPR} (computed only for 3D-aware methods). To provide a common metric to 2D and 3D-aware image generation methods that cannot synthesize videos, we use the Frechet Image Distance (FID) \citep{Heusel2017NIPS}. The individual frames for FID scores are randomly selected from all generated video frames.
\rebuttaltmlr{Note that the checkpoints used for the results of StyleGAN2, VideoGPT, MoCoGAN, and MoCoGAN-HD on FaceForensics were not released by the StyleGAN-V authors. Hence, we are not able to compute the ACD and CPBD metrics for these methods. We can not compute ACD for StyleNeRF, as it is a static method.}

\begin{figure*}[t]
    \centering
    \includegraphics[width=\textwidth]{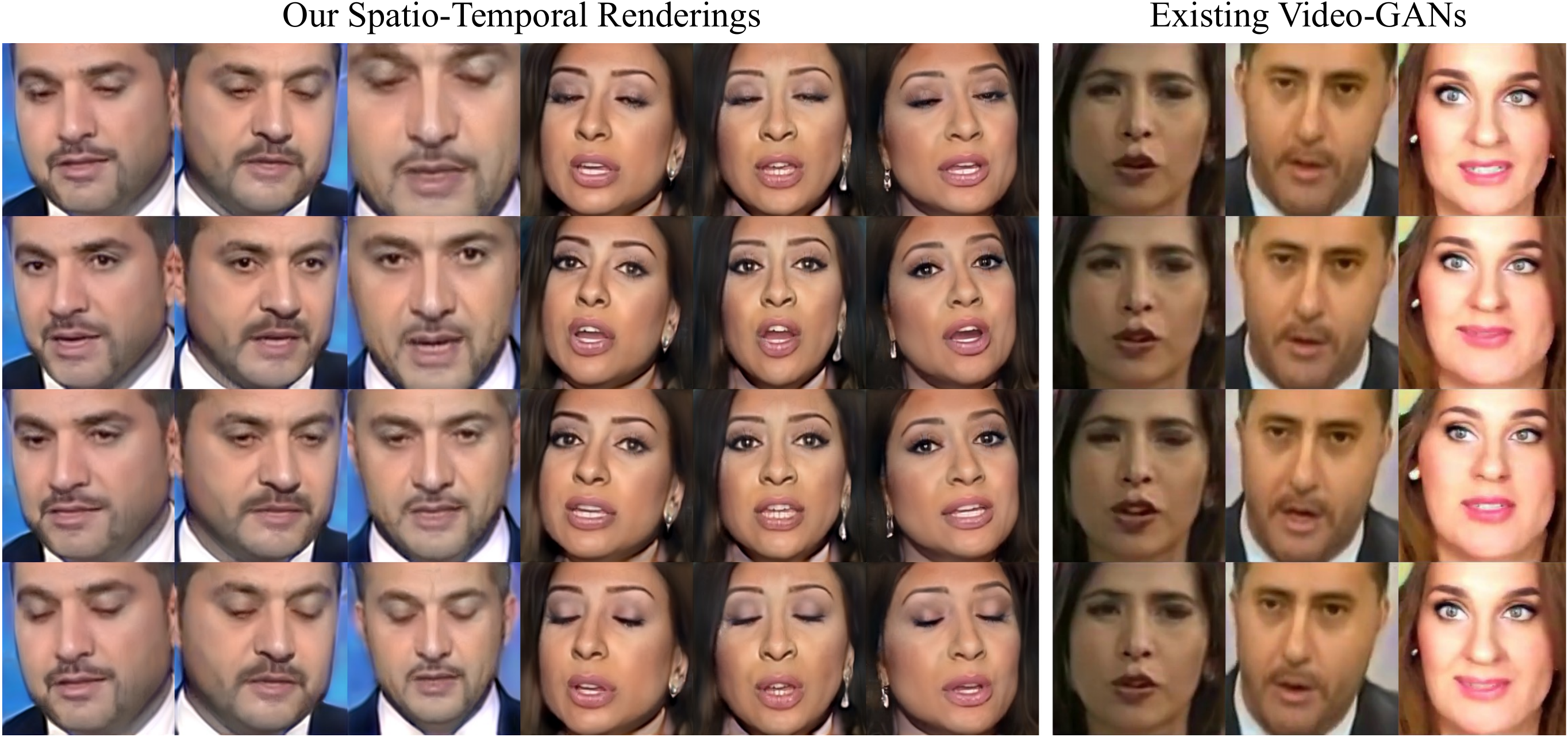}
    \vspace{-1.2em}
    \caption{{\bf{Qualitative results on FaceForensics Dataset.}} The first 6 columns visualize two 4D fields sampled from of our model trained on FaceForensics rendered from various spatio-temporal snapshots. Note the high quality visuals and motions across diverse viewpoints. The last three columns show the video samples of MoCoGAN-HD \citep{Tian2021ICLR}, DIGAN \citep{Yu2022ICLR}, and StyleGAN-V \citep{Skorokhodov2022CVPR}, in that order.
    }\label{fig: faceforensics}
    \vspace{-1.0em}
\end{figure*}

\begin{figure*}[t]
    \centering
    \includegraphics[width=\textwidth,height=6cm]{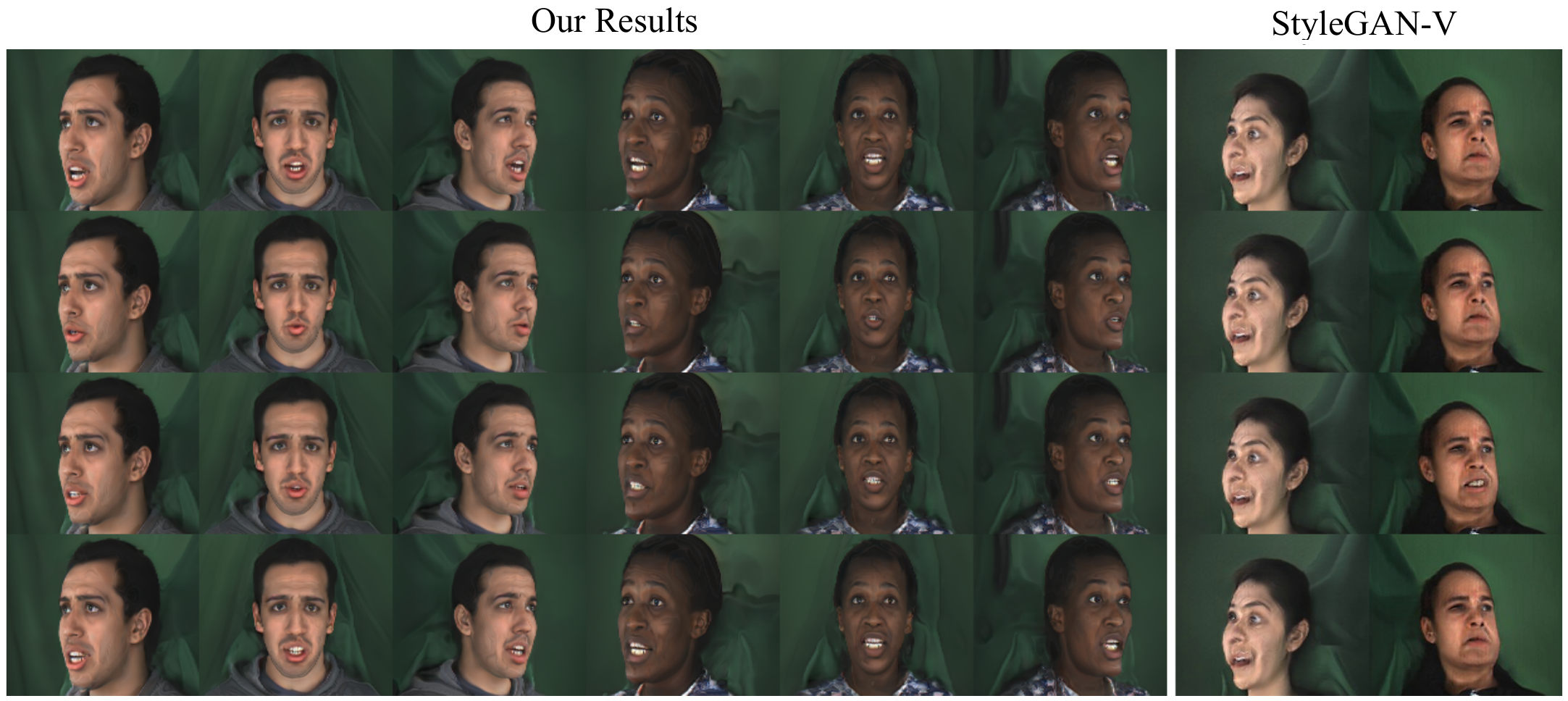}
    \vspace{-1.5em}
    \caption{{\bf{Results on the MEAD Dataset.}} The first six columns  show the spatio-temporal renderings of our 4D GAN, where the vertical axis indicates the progress in time. The two right-most columns show generated videos from StyleGAN-V \citep{Skorokhodov2022CVPR}. Zoom-in to inspect details.}\label{fig: mead}
    \vspace{-0.6em}
\end{figure*}

        


\begin{table}[t!]
    \centering
    \begin{tabularx}{0.8\linewidth}{l|lccccc}
        \toprule
        \multirow{1}{*}{Type} & \multirow{1}{*}{Method} & FVD ($\downarrow$) & FID ($\downarrow$) & ACD ($\downarrow$) & CPBD ($\uparrow$) & ID ($\uparrow$)\\
        \midrule
        \multirow{1}{*}{2D Image} & StyleGAN2  & - & 8.4 & - & - & -\\
        \midrule
        \multirow{5}{*}{2D Video}     & VideoGPT  & 185.9& 22.7 & - & - & -\\    & MoCoGAN  & 124.7& 24.0 & - & - & -\\
        & MoCoGAN-HD & 111.8 & 7.1 & - & - & -\\
        
        & DIGAN  & 62.5 & 19.1& 1.09 & 0.171 & -\\
        & StyleGAN-V  & 47.4 & 9.4& 1.11 & 0.155 & - \\

        \midrule
        \multirow{1}{*}{3D Static}         & StyleNeRF & - & 15.3& - & 0.181 & 0.812\\
        \midrule
        \multirow{1}{*}{3D Video} & Ours & \todo{68.7} & \todo{13.7}& 0.965 & 0.196 & 0.861\\
        \bottomrule
    \end{tabularx}
    \caption{{\bf{Quantitative Results on FaceForensics.}} We report metrics for all methods at $256^2$ pixel resolution. For FVD and FID of 2D image and video methods, we use the numbers reported in StyleGAN-V. We omit ACD and CPBD metrics for methods whose checkpoints are not publicly available. View-consistent identity (ID) is only computed for 3D-aware methods.}
    \label{tab:vid_gen_sota_ff}
\end{table}

    \begin{table}[!t]
    \centering
\begin{minipage}[t]{0.52\linewidth}\centering
    \label{tab:vid_gen_sota_mead}
    \begin{tabularx}{0.99\linewidth}{@{}X@{}ccc@{}}
        \toprule
        Method & FVD ($\downarrow$) & ACD($\downarrow$) & CBPD($\uparrow$)\\
        \midrule
        StyleGAN-V  & \todo{109.3} & 0.136 &0.509 \\
        Ours & \todo{55.4} & 0.060 &0.469 \\
        \bottomrule
    \end{tabularx}
     \caption{{\bf{Results on MEAD Dataset.}}}
\end{minipage}\hfill%
\begin{minipage}[t]{0.47\linewidth}
\centering
\label{tab:vid_gen_sota_taichi}
    \begin{tabularx}{0.99\linewidth}{@{}X@{}ccc@{}}
        \toprule
        Method & FVD($\downarrow$) & ACD($\downarrow$) & CBPD($\uparrow$) \\
        \midrule
        DIGAN  & \todo{151.7} & 0.537 & 0.672\\
        Ours & \todo{158.3} & 0.552&0.632 \\
        \bottomrule
    \end{tabularx}
    \caption{{\bf{Results on TaiChi Dataset.}}}
\end{minipage}
\end{table}

\begin{figure*}[t]
    \centering
    \includegraphics[width=\textwidth]{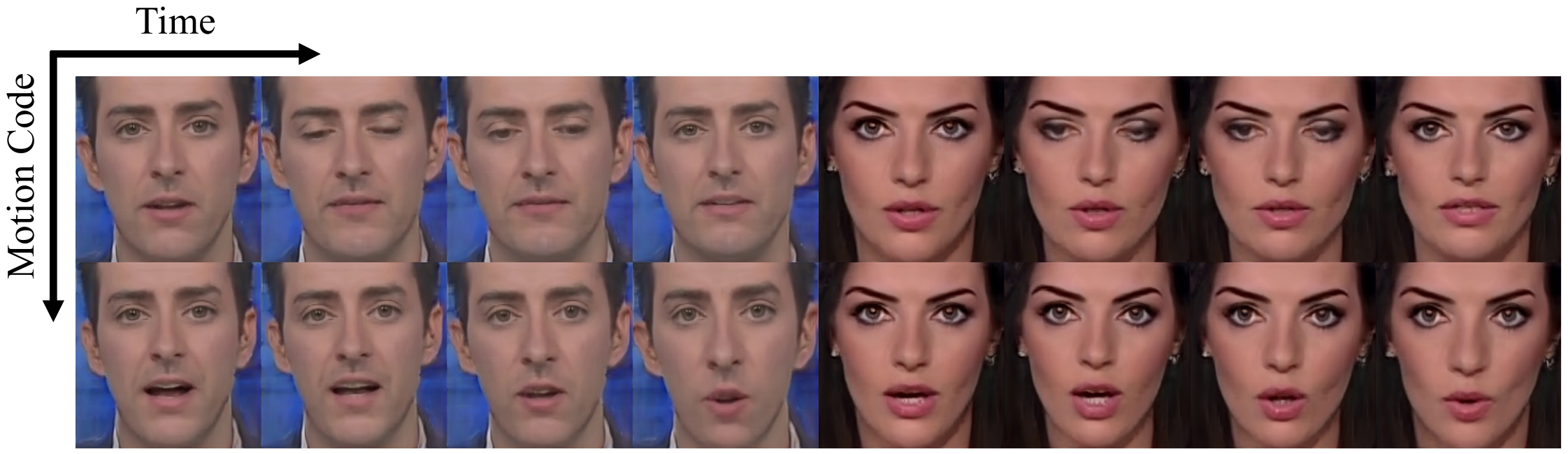}
    \vspace{-0.3em}
    \caption{{\bf{Motion and content decomposition.}}
    The two rows show respective video sequences that share the same 3D content latent vector applied to different motion vectors. Here we show motions with four time steps for two different identities. Note the difference in motions while the identities of the person appear unchanged.}\label{fig: decomposition}
    \vspace{-1.2em}
\end{figure*}

\vspace{-0.2em}
\subsection{Main Results}
\vspace{-0.2em}
In this section, we provide experimental evidence that our model is able to learn a distribution of 3D-aware videos that permit spatio-temporal control. In particular, in Figures~\ref{fig:teaser},~\ref{fig: taichi1},~\ref{fig: faceforensics}, and~\ref{fig: mead}, we visualize snapshots of sampled 4D scenes rendered from various time steps and camera angles across various datasets. Moreover, we conduct both qualitative and quantitative comparisons  against the strongest video GAN baselines on the all datasets. For each method, we use video length of 16 frames and use image resolution of $256^2$ for FaceForensics and MEAD, and $128^2$ for TaiChi dataset. 

FaceForensics is one of the most widely used dataset in the video generation literature, and thus we measure the scores across a wide range of methods, including: 2D image \citep{Karras2020CVPR} and video generation works \citep{Yu2022ICLR, Skorokhodov2022CVPR,Tulyakov2018CVPR,Tian2021ICLR,Yan2021ARXIV}, 

\begin{wraptable}{r}{8cm}
\vspace{-0.em}
\centering
    \vspace{-0.0em}
    \begin{tabularx}{0.8\linewidth}{@{}X@{}c@{}}
        \toprule
        Method & FVD  \\
        \midrule
        Ours & \todo{158.3} \\
        \midrule
        w/o Background NeRF & \todo{203.7}  \\
        w/ Static and Dynamic Separation & \todo{207.6}  \\
        w/o Motion Generator & \todo{166.3}  \\
        w/ Positional Time Encoding  & \todo{175.4}  \\
        w/o Image Discriminator & \rebuttal{234.3}  \\
        \midrule
    \end{tabularx}
    \caption{{\bf{Ablation studies on TaiChi Dataset.}}}
    \label{tab:ablation_taichi}
\vspace{-1em}
\end{wraptable}

and a 3D generation method \citep{Gu2022ICLR}.
The qualitative results in Fig.~\ref{fig: faceforensics} show our model's ability to generate 4D visual effects, displaying the spatiotemporal renderings and the physically-based zoom-in effects. 
The quantitative results in Table~\ref{tab:vid_gen_sota_ff} confirm that our generated imagery are of competitive quality against the strongest 2D and 3D baselines. Note that the metrics for the 2D image and video methods are copied from the reports of \cite{Skorokhodov2022CVPR}. We train StyleNeRF \citep{Gu2022ICLR} on FaceForensics from scratch. 

Fig.~\ref{fig: taichi1} presents sampled scenes from our TaiChi-trained model with varying time and viewpoints, showcasing our 4D learning on a challenging setup with complex motions and backgrounds. When compared with one of the latest 2D video generation methods, DIGAN \citep{Yu2022ICLR}, our method synthesizes frames with competitive visual fidelity while introducing another degree of freedom. Note that we used the provided pre-trained model to recompute their score following the FVD protocol used by \cite{Skorokhodov2022CVPR}.  

The MEAD dataset contains videos of faces taken from a wide range of viewpoints. We hypothesize that, for the 2D-based methods, a significant amount of expressive power will be wasted for redundantly modeling the diverse views in 2D, while the 3D approach would only need to learn a shared representation across views. The results, shown in Fig.~\ref{fig: mead}, indeed suggest higher visual quality for our 4D approach, while allowing rendering from 7 different viewpoints. Our quantitative analysis in Tab.~\ref{tab:vid_gen_sota_mead}
supports our observations -- our FVD is noticeably better than that of StyleGAN-V \citep{Skorokhodov2022CVPR}. Note that the pre-trained model provided by \cite{Skorokhodov2022CVPR} was trained only on frontal views, so we included non-frontal faces and re-trained. We omit comparing against other video methods on MEAD that are already compared in \cite{Skorokhodov2022CVPR}.
Finally, we demonstrate that the use of two independently sampled latent vectors to modulate motion and content makes these two components separable.  Fig.~\ref{fig: decomposition} showcases such decomposition by applying two different motion codes to the same content code. 
We refer the reader to our supplementary for additional results and videos to fully appreciate 4D renderings.

\vspace{-0.2em}
\subsection{Ablation}
\label{sec:exp_ablations}

We conduct ablation studies to gauge how the design choices of our algorithm affect the quality of 4D scene generations. We use TaiChi dataset, as its diverse scenes and motions help us identify the effects of each component. \cref{tab:ablation_taichi} summarizes the numerical results, \rebuttal{for further details see Sec. \ref{sec:ablations_appendix} in the appendix.}
\vspace{-0.2em}
\paragraph{Background NeRF}
We analyze the importance of the background NeRF based on the inverse sphere parametrization of NeRF++ \citep{Zhang2020ARXIV} used for capturing unbounbed scenes. We observe that using the additional background NeRF  is critical for the model to  disentangle the dynamic and static parts in the video, leading to less motion artifacts and more static backgrounds.

\paragraph{Static and Dynamic Separation}
\vspace{-0.2em}
We forgo the use of the inverse sphere background NeRF of \cite{Zhang2020ARXIV} and instead decompose the scene with static and dynamic NeRFs, following \cite{Gao2021ICCV}. 
We observe that this setup hurts the generation quality, suggesting that the background parameterization of NeRF++ \citep{Zhang2020ARXIV} is critical for training a 4D generative model.

\paragraph{Motion Generator}
\vspace{-0.2em}
Here we omit the use of motion vector and its mapping network, and instead pass the time value directly to the foreground NeRF. The use of motion network  improves the output quality and induces useful decomposition of motion and 3D content.




\paragraph{Positional Time Encoding}
\vspace{-0.2em}
Our current motion mapping MLP is conditioned on the raw time value.  We observe that applying positional encoding on the Fourier features leads to repetitive and unnatural motions even when tuning the frequencies of the Fourier features. We leave applying more complicated positional encoding of time as future work.


\paragraph{Image Discriminator}
\vspace{-0.2em}
Moreover, we demonstrate that adding a discriminator that specializes on single-frame discrimination improves quality compared to using only a time-aware discriminator.

\vspace{-0.2em}
\section{Limitations \& Discussions}
\vspace{-0.2em}
Our approach models the entire dynamic foreground with a single latent vector, which limits the expressive capacity of our generative model. Learning with more number of independent latent vectors, as tried in \citep{Niemeyer2021CVPR,Hudson2021ICMLR}, could promote  handling of multi-objected scenes with more complex motions. Furthermore, our training scheme assumes static camera and dynamic scenes, so it cannot handle videos taken from a moving camera. Modeling plausible camera paths is an interesting but less explored problem.
Being the first 4D GAN approach, our method can only generate short video sequences (16 time steps, following \citet{Yan2021ARXIV}, \citet{Tian2021ICLR}, and \citet{Yu2022ICLR}). We leave applying recent progress \citep{Skorokhodov2022CVPR} in generating longer videos as future work.
Moreover, our model uses 2D upsampling layers to efficiently render higher resolution images, which breaks multi-view consistency. One possible solution is patch-based video training as recently investigated in the 3D-aware image generation works EpiGRAF \citep{skorokhodov2022epigraf}, SinGRAF \citep{son2022singraf}, and 3DGP \citep{skorokhodov20233d}.
Finally, we model the scene motion as a purely statistical phenomenon,
without explicitly considering physics, causality, semantics, and entity-to-entity interactions. These topics remain important challenges, which we continue to explore.

\vspace{-0.2em}
\section{Conclusions}
\label{sec:conclusions}
\vspace{-0.2em}
In this work, we introduced the first 4D generative model that synthesizes realistic 3D-aware videos, supervised only from a set of unstructured 2D videos. Our proposed model combines the benefits of video and 3D generative models to enable new visual effects that involve spatio-temporal renderings, while at the same time maintaining the visual fidelity. The resulting latent space of rich, decomposable motion and 3D content sets up the foundation to exciting new avenue of research towards interactive content generation and editing that requires knowledge of underlying 3D structures and motion priors.



\section{Ethics Statement}
\subsection{Potential Misuse}
We condemn the use of neural rendering technologies towards generating realistic fake content to harm specific entities or to spread misinformation, and we support future research on preventing such unintended applications. 

While our method can plausibly synthesize moving human heads and bodies, the technology in the current form does not pose significant threat in enabling the spread of misinformation. We highlight that our 4D GAN is an unconditional generative model. This means that the current method can only generate {\em random} identities and motions and thus cannot be used for targeting specific individuals for misinformation (e.g., creating a video of a particular politician). Similarly, being an unconditional model, we cannot control a person with an user-desired motion, e.g., following a text prompt. 

However, we acknowledge the possible development of future 4D GAN research that trains conditional models that are able to synthesize videos of given individuals. We note that there are various existing technologies to spot neural network-generated imagery with surprising level of accuracy \citep{cozzolino2021id,yang2021mtd,de2020deepfake,guera2018deepfake,kumar2020detecting,wang2020cnn}. Among them, \cite{wang2020cnn} suggests that the key of training such a detector is a realistic generator, which can provide ample amount of training data for fake content detection. In this context, we believe that our unconditional 4D GAN can be effectively used to fight against the potential misuse of AI-generated videos by generating such training data, because it can sample realistic videos from diverse identities, motions, and camera viewpoints. 

\subsection{Privacy Issues}
We note readers that our experiments only used publicly available datasets that contain videos that belong to public domains. Moreover, we emphasize that all of our figures and videos across the main paper and the supplementary materials contain {\em synthetic} content. Therefore, {\em all} of the human faces or bodies shown in this work do not involve any privacy concerns.

\subsection{Diversity and Potential Discrimination}
We have shown in our main paper and the supplementary that our trained model on the FaceForensics and MEAD datasets are able to generate diverse human faces across gender and ethnicity. We acknowledge that our generation results might not cover diversity of some human traits, e.g., weights, disabilities, etc. Note the TaiChi dataset primarily contains videos of certain ethnicity due to the skewed popularity of the activity.

\section{Reproducibility Statement}
We provide detailed implementation-related information of our algorithms in Sec. C of the supplementary document. We will release the source code for training and testing our algorithms upon acceptance. For experiments. we describe our datasets (along with how we processed them) and metrics in Sec.~\ref{Sec: experiment setups} and provide in-depth details of our experiments in Sec. D of the supplementary document to further improve the reproducibility.
\section{Acknowledgements}

This project was supported in part by ARL grant W911NF-21-2-0104, a Vannevar Bush Faculty Fellowship, a
gift from the Adobe Corporation, a PECASE by the ARO, NSF award
1839974, Stanford HAI, and a Samsung GRO.
Despoina Paschalidou was supported by the Swiss National Science Foundation under grant number P500PT\_206946.

\bibliography{bibliography_long,bibliography,bibliography_custom}
\bibliographystyle{tmlr}

\appendix
\section{Video Results}

We urge readers to view our video results by opening \url{https://sherwinbahmani.github.io/3dvidgen}.
We provide video results on the FaceForensics \citep{Rossler2019ICCV}, MEAD \citep{Wang2020ECCV}, TaiChi \citep{Siarohin2019NIPS}, and SkyTimelapse \citep{xiong2018learning}. In particular,
we show the generated videos for all datasets from various camera
viewpoints in order to showcase the ability of our model to learn a distribution of 3D-aware videos, which we highly encourage readers to view.

Specifically, for the case of FaceForensics, we show generated videos with a forward-facing camera (see Ours with Forward-facing Camera), with a camera that rotates along the yaw-axis (see Ours with Rotating Camera) and with a forward-facing camera that moves away from the depicted individual, thus creating a zoom-out effect, (see Ours with Forward-facing Camera and Zoom Effect).
Moreover, we also provide generated videos using a variant of our model that uses a pre-trained generator on FFHQ \citep{Karras2019CVPR}, as discussed in the main submission. This variant of our model allows a wider range of viewpoint control than our original model. 
Particularly, for this variant of our model, we also show examples of generated videos, while we rotate the camera along both yaw and pitch axes, where the fine-tuned model generates realistically looking videos.
In addition, we show "Motion and Content Decomposition" examples, where we show the ability of our approach to control shape and motion separately; i.e., we can create videos illustrating the same human performing  different motions, and vice versa.
We also provide example videos of prior work including MoCoGAN-HD \citep{Tian2021ICLR}, DIGAN \citep{Yu2022ICLR} and StyleGAN-V \citep{Skorokhodov2022CVPR}.
We note that our generated videos are of comparable quality against those of the state-of-the-art video generation methods \citep{Yu2022ICLR, Skorokhodov2022CVPR}, 
while at the same time permitting control on the camera viewpoint, e.g., zoom-out to reveal new content or rotate the camera around, which is not possible for the latest 2D video methods.

Similarly, we also showcase examples of our generated videos on the MEAD \citep{Wang2020ECCV} dataset using a similar setup. We observe that in comparison to StyleGAN-V \citep{Skorokhodov2022CVPR} our generations have significantly fewer visual artifacts (the faces of \cite{Skorokhodov2022CVPR} appear uncanny), while at the same time our generated videos from different camera viewpoints (see Ours with Different Camera Positions) are consistently plausible. 

We also show examples of generated videos on the TaiChi \citep{Siarohin2019NIPS} dataset using a similar setup. In addition, we also consider two more setups, where we rotate the camera along the yaw-axis while having a static human (see Ours with Rotating Camera and Static Motion) and a moving human (see Ours with Rotating Camera and Dynamic Motion). Also for these scenarios, the quality of our generated videos are comparable to that of DIGAN \citep{Yu2022ICLR} that does not allow viewpoint control.

Finally, we show samples for the SkyTimelapse \citep{xiong2018learning} dataset. The dataset is in contrast with the other three datasets, as its videos often contain multiple objects or entities. We provide videos rendered from a fixed camera, and a rotating camera along the yaw-axis with and without scene dynamics (Ours with Rotating Camera along First Axis and Static Motion, and Ours with Rotating Camera along First Axis and Dynamic Motion). Similarly, we rotate the camera along the pitch-axis with and without scene motions (Ours with Rotating Camera along Second Axis and Static Motion, and Ours with Rotating Camera along Second Axis and Dynamic Motion). Note that we only model rotation of a camera located at the origin. A careful modeling of camera distribution for such a large-scale scene dataset is out of scope of this work, and thus we omit an in-depth analysis.

\section{Visualizing Latent Interpolations and Depth Maps}
Our 4D GAN with decomposed content and motion latent spaces allow interpolation of content with fixed motion and vice versa. Moreover, our sampled neural fields can be used to obtain depth maps given the 3D nature of our representation.

\begin{figure*}[t]
    \centering
    \includegraphics[scale=0.19]{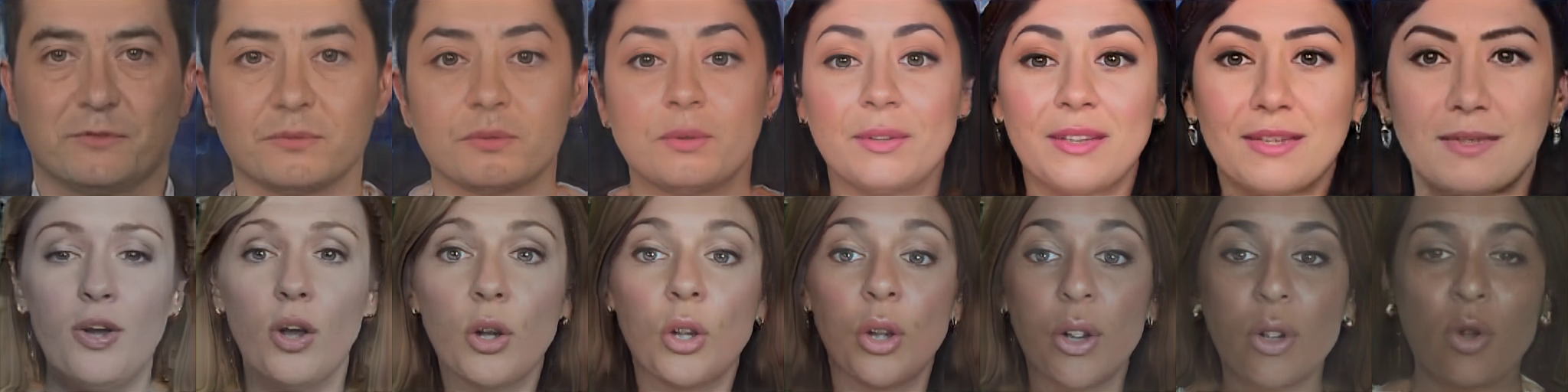}
    \caption{{\bf{Interpolation in the content latent space.}} We visualize, for each row, examples of linearly interpolating between two sampled latents in the content space with fixed motion vectors and camera viewpoints. Note the smooth and plausible transition of face appearances. }\label{fig:content_interpolation}
\end{figure*}

\begin{figure*}[t]
\centering
    \vspace{1.5em}
    \includegraphics[scale=0.19]{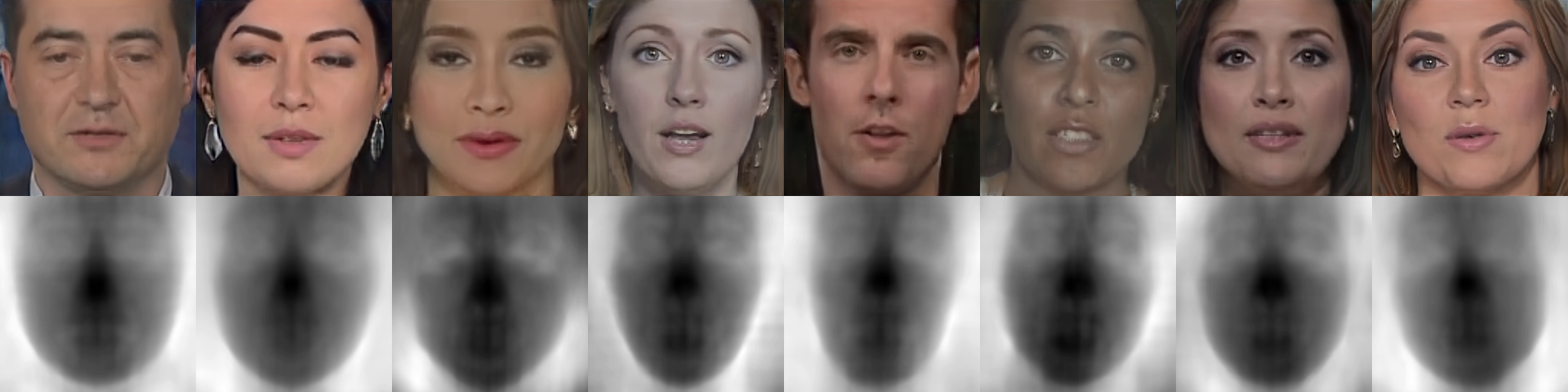}
    \caption{{\bf{Depth visualizations (FaceForensics).}} We show depth maps (second row) obtained by volume rendering the sampled 4D fields at a given time step. The first row shows the corresponding RGB rendering of the same 4D fields. Here the depth is defined as the expected ray termination distance.}\label{fig:faceforensics_depth}
\end{figure*}

\begin{figure}[t]
\centering
    \includegraphics[scale=0.38]{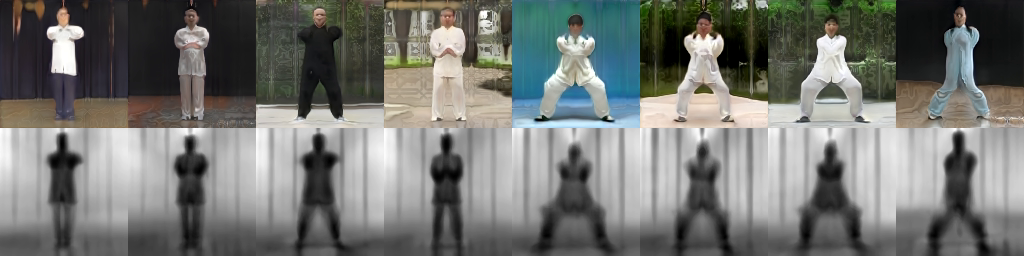}
    \rebuttaltmlr{\caption{{\bf{Depth visualizations (TaiChi).}} We show depth maps (second row) obtained by volume rendering the sampled 4D fields at a given time step. The first row shows the corresponding RGB rendering of the same 4D fields. Here the depth is defined as the expected ray termination distance.}\label{fig:taichi_depth}}
\end{figure}

As shown in Fig. \ref{fig:content_interpolation}, linearly interpolating between two sampled content vectors maps to smooth, plausible interpolation of content appearance in the 4D fields.
Similarly, we fix the content vector and apply interpolated motion vectors. Such visualization is best viewed as videos, so we refer readers to the supplementary website \texttt{supp.html} (see Ours with Motion Interpolation).
The video results show that the interpolation in the motion latent space leads to smooth transition of motions.
Finally, in Fig. \ref{fig:faceforensics_depth} and Fig. \ref{fig:taichi_depth}, we visualize example depth maps obtained via volume rendering our sampled neural fields.

\section{Implementation Details}
\label{sec:app_implementation}

\subsection{Architecture and Training Details}
 The 3D content code, motion code and style vector dimensions are all set to 512. Our motion generator (see \figref{fig:model_architecture}) is implemented as an MLP with three fully connected (FC) layers and Leaky ReLU activations. 
 The time step is repeated across the channel dimension and multiplied with the output of the first fully connected layer of the motion generator. We set the motion code and hidden dimension of the motion generator to 512, while the output dimension is 128. The output of the motion generator is then added to the output of the first FG NeRF Block, which also is a 128-dimensional representation. \rebuttaltmlr{Besides the virtual camera setup and the motion generator dimensions, we adopt all hyperparameters determined in StyleNeRF \citep{Gu2022ICLR}. We set the motion generator hyperparameters empirically following previous GAN works. We believe that hyperparameter grid search and increasing the model dimensions could further boost the quality of our results, this, however, requires more computational resources. We analyze the virtual camera setup in the subsequent section.}
 
 Our foreground and background NeRF are modeled as MLPs (with Leaky ReLU activations) with 8 and 4 FC layers that each contain 128 and 64 hidden units, respectively. We use 10 frequency bands to map the positional input of the foreground background NeRF to the fourier features \citep{Mildenhall2020ECCV}. We do not apply positional encoding to the time input.  We follow the implementation of StyleNeRF \citep{Gu2022ICLR} for the the 2D ray aggregation and upsampling block (see \figref{fig:model_architecture}) and the volume rendering process.
 Both the image and video discriminator follow the architecture of StyleGAN2 \citep{Karras2020CVPR} with hidden dimensions of 512, and the input channels being 3 and 7, respectively. We apply the Differentiable Augmentation technique \citep{Zhao2020NIPS} with all augmentations except CutOut, to prevent the discriminators from overfitting to the relatively small video datasets. In contrast to StyleNeRF \citep{Gu2022ICLR}, we do not use progressive-growing training \citep{Karras2018ICLR} but directly train on the final image resolution, as we did not observe any change in visual quality.  
 
For both the generator and discriminator, we use the Adam optimizer \citep{Kingma2015ICLR} with a learning rate of 0.0025, $\beta_1 = 0$, $\beta_2 = 0.99$ and $\epsilon = 10^{-8}$. We follow the setup of StyleGAN2 \citep{Karras2020CVPR} to use 8 fully connected layer content mapping network and apply 100$\times$ lower learning rate compared to that of the main generator layers.
 For our objective function (Eq.~\ref{eq: objective}), we set $\lambda_1 = 0.5$ and $\lambda_2 = 0.2$. We use 16 samples for the NeRF path regularization \citep{Gu2022ICLR}.

\subsection{Virtual Camera Setup}
For the three main datasets we empirically set the virtual camera on the surface of unit sphere and parameterize the camera viewpoint distribution with pitch and yaw angles. The standard deviation for pitch sampling is 0.15 for all three datasets. For yaw sampling the standard deviation is 0.3, 0.3, and 0.8 for FaceForensics, MEAD, TaiChi. The field-of-view of the camera is set to be 18 degrees.

For the SkyTimelapse dataset we do not sample on a sphere, but place the camera at the origin and make the camera look outwards.  We uniformly sample a point on a hemisphere and set the camera to look towards the direction. The field-of-view is set to be 80 degrees. Note that this setup only models rotation of the camera. We leave a more complicated camera sampling method as future work.

\subsection{\rebuttaltmlr{Motion and Content Disentanglement}}

\rebuttaltmlr{
The motion and content are mainly disentangled because of our two-frame discrimination in combination with the different sampled latent codes. While the content code is fixed across the two frames, the intermediate motion code computed from the sampled motion code and variable timesteps are not constant across the two frames. Hence, the FG network receives two different motion features for the same content during training. This inductive bias allows the model to disentangle content from motion. Note that we mainly follow the 2D video generator MoCoGAN \citep{Tulyakov2018CVPR}, which disentangles motion and content in a similar fashion and sets a default approach for motion and content decomposition in video generation for most of the follow-up works. In early experiments, we observed slightly more content inconsistencies when injecting the content at deeper layers, hence we inject the motion directly after the first layer. We also choose the output of the first layer and not the input of the first layer as this allows us to use the addition operation by first mapping the number of channels to a common channel number for motion generator output and FG MLP. Otherwise we have to use, e.g., concatenation which does not allow the easy pre-training technique by disabling motion in the sum as discussed in Sec. \ref{sec:method_training}.}

\section{Experiment Details}

\subsection{Evaluations}
\begin{figure*}[t]
    \centering
    \includegraphics[scale=0.2]{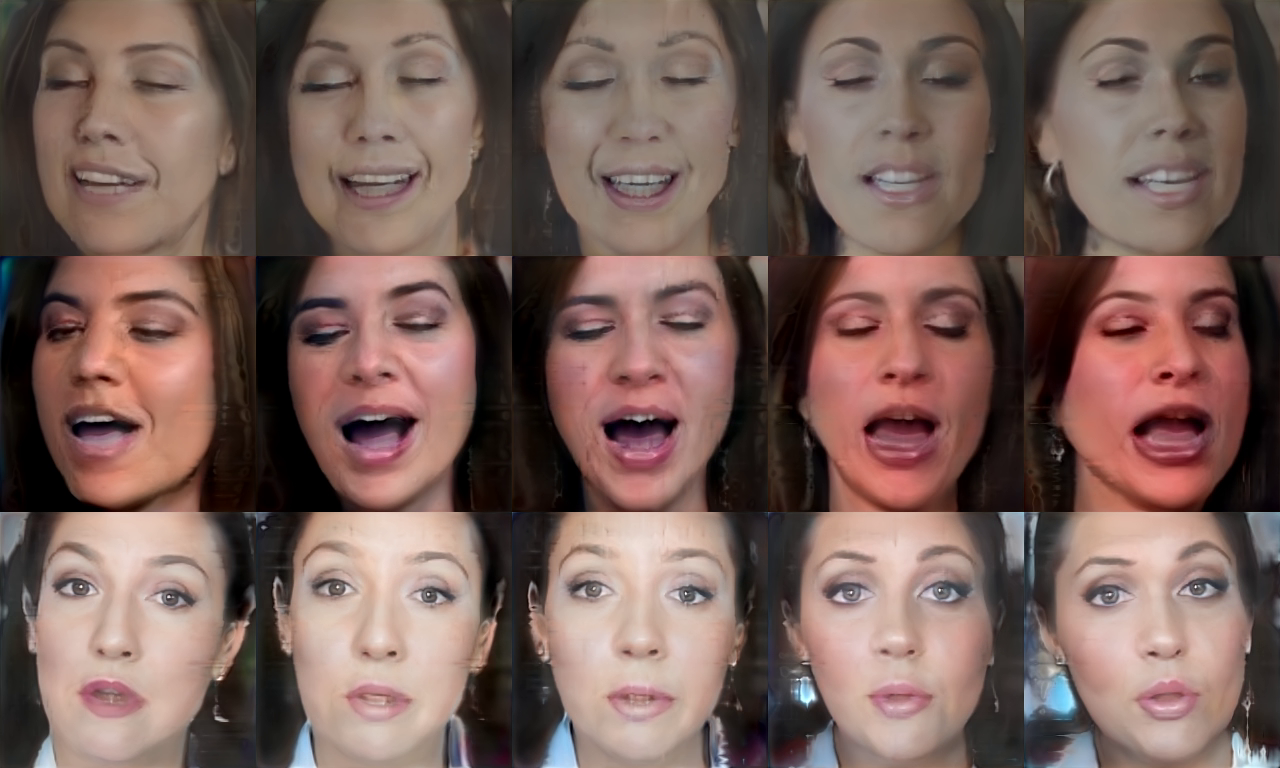}
    \vspace{-0.5em}
    \caption{{\bf{StyleNeRF results.}} We show qualitative results of StyleNeRF model trained on FaceForensics, rendered from five different cameras (columns) for three identities (rows). }
    \label{fig:stylenerf_grid}
    \vspace{-0.0em}
\end{figure*}

We used the code and evaluation protocol of StyleGAN-V \citep{Skorokhodov2022CVPR} for computing FVD \citep{Unterthiner2018ARXICV} and FID \citep{Heusel2017NIPS}. The FVD protocol requires 2048 16-frame videos, while the FID score uses 50K images. For MEAD dataset \citep{Wang2020ECCV}, we re-trained StyleGAN-V \citep{Skorokhodov2022CVPR} using their official code as the authors only provided results on the front view videos of MEAD at $1024^2$ resolution. We randomly choose 10,000 videos across all viewpoints, including non-frontal views and follow the identical training setup provided by \cite{Skorokhodov2022CVPR} to process 25,000K images with batch size 64. 
For TaiChi dataset,  we use the officially provided checkpoint of DIGAN \citep{Yu2022ICLR} to evaluate their model with the new FVD protocol \citep{Skorokhodov2022CVPR}. For FaceForensics dataset, we use the reported numbers provided by the StyleGAN-V \citep{Skorokhodov2022CVPR} authors for all models in Table \ref{tab:vid_gen_sota_ff}, except for StyleNeRF \citep{Gu2022ICLR} and our model, which we train from scratch. We provide sampled 3D models of StyleNeRF trained on FaceForensics, rendered from a horizontally moving camera, as shown in Fig. \ref{fig:stylenerf_grid}.


We train our model and StyleNeRF using 4 NVIDIA V100 GPUs. For our approach we train for a maximum of 3,000K images, which takes two days at $256^2$ resolution. For StyleNeRF we abort the training after 15,000K images due to the FID diverging after 11,400K images.
It is generally true that 2D models like StyleGAN-V \citep{Skorokhodov2022CVPR} have a significantly faster throughput, as there is no costly volume rendering. However, we observe that our 4D model converges at much lower iterations, already converging after 2,500K processed images, while some of the 2D video models such as VideoGPT \citep{Yan2021ARXIV} does not converge even at 25,000K processed images. We hypothesize that the extraordinary fast convergence of our model is due to the explicit disentanglement of 3D content, camera viewpoints, and motions. 


\subsection{Statistical Reproducibility}
For computing the FVD, we follow the protocol of StyleGAN-V \citep{Skorokhodov2022CVPR}, which aims to reduce the score variations significantly compared to the original protocol \citep{Unterthiner2018ARXICV}. 
 Nevertheless, in Table~\ref{tab:standard_deviations} we report the relative standard deviations after evaluating our three main dataset results for 10 rounds with the FVD protocol \citep{Skorokhodov2022CVPR}. We observe that the standard deviation is rather small across all three datasets, as reported in the extensive analysis of \cite{Skorokhodov2022CVPR}.

 \begin{table}
 \centering
 \begin{tabularx}{0.6\linewidth}{@{}X@{}c@{}}
        \toprule
        Dataset & FVD Relative Standard Deviation  \\
        \midrule
        FaceForensics & 2.62\% \\
        TaiChi & 1.65\% \\
        MEAD & 2.17\% \\
        \midrule
    \end{tabularx}
    \caption{Relative standard deviation for the FVD metric on FaceForensics, TaiChi, and MEAD datasets, computed as percentage of standard deviation with respect to the mean.}
    \label{tab:standard_deviations}  
\end{table}


\subsection{\rebuttaltmlr{3D Consistency Evaluation}}

\rebuttaltmlr{In Table~\ref{tab:chamfer_distance} we further report an evaluation of the underlying 3D model using the chamfer distance (CD) metric. Specifically, we follow the protocol of \citet{or2022stylesdf} and evaluate chamfer distance between the front pose and a randomly sampled side pose on 1000 generated FaceForensics samples. We compare our results to StyleNeRF, as it is the only 3D-aware baseline. We observe a lower chamfer distance, which aligns with our previous results that we can extend a 3D-aware image generator in the temporal domain to generate 3D videos without sacrificing any visual quality or geometric consistency.}

 \begin{table}[!t]
 \centering
 \rebuttaltmlr{\begin{tabularx}{0.3\linewidth}{@{}X@{}c@{}}
        \toprule
        Method & CD ($\downarrow$) \\
        \midrule
        StyleNeRF & 1.27 \\
        Ours & 1.03 \\
        \midrule
    \end{tabularx}}
    \caption{\rebuttaltmlr{Chamfer distance evaluation on FaceForensics between StyleNeRF and our approach, following the protocol of \citet{or2022stylesdf}.}}
    \label{tab:chamfer_distance}  
\end{table}

\begin{figure*}[t]
    \centering
    \includegraphics[width=\textwidth]{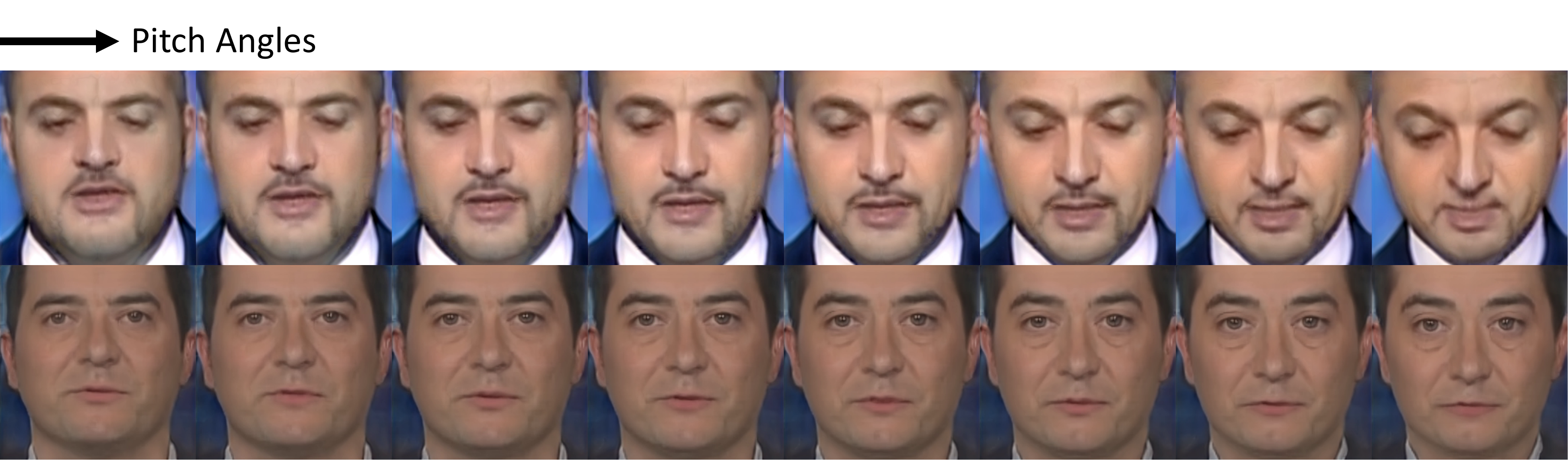}
    \vspace{-0.5em}
    \caption{{\bf{Comparing the range of pitch control.}} We show that our 4D model trained solely on the FaceForensics \citep{Rossler2019ICCV} dataset (top row) has a narrow range of pitch where high quality renderings can be produced, likely due to the lack of diverse training views. Note that the upper and bottom parts of the head becomes unnaturally large at both extremes. Fine-tuning a model pre-trained on FFHQ \citep{Karras2019CVPR} image dataset (bottom row), which features more diverse view angles, results in a wider range of views that generate plausible images.}
    \label{fig:pitch}
\end{figure*}

 \begin{figure*}[t]
    \centering
    \includegraphics[scale=0.2]{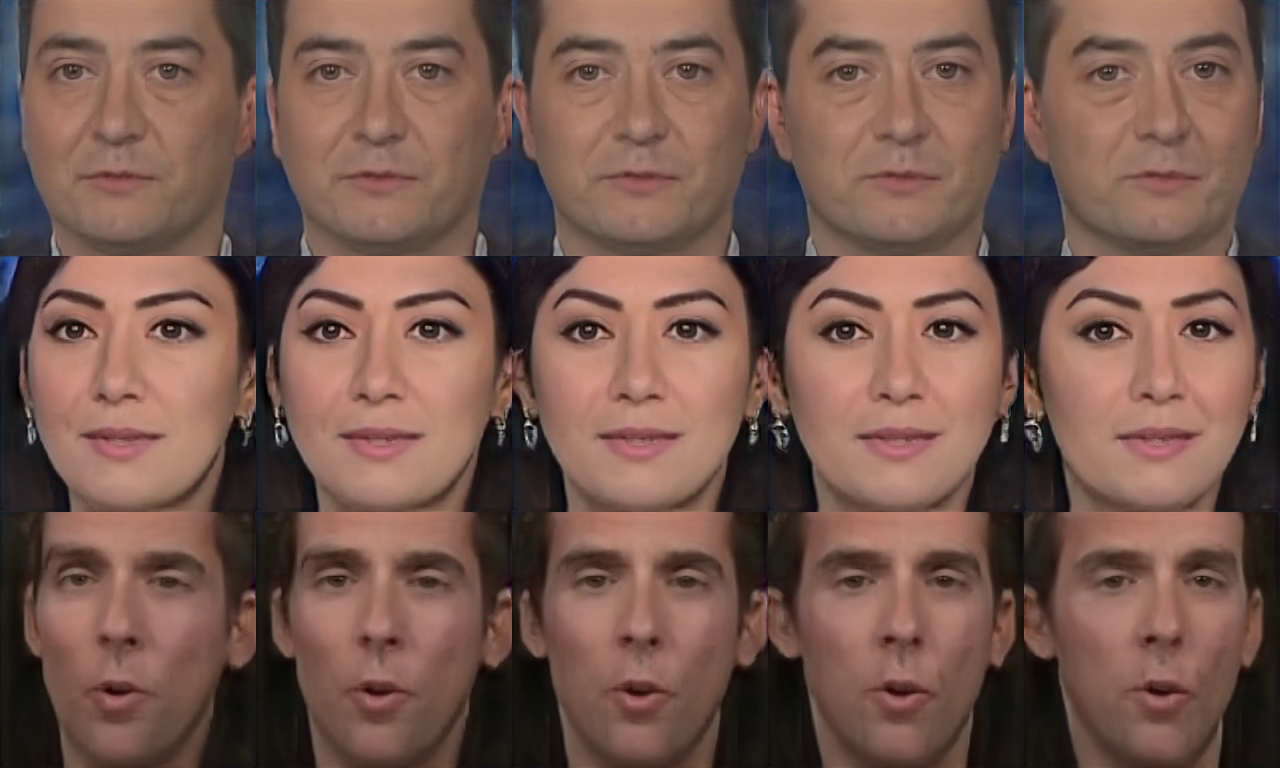}
    \vspace{-0.5em}
    \rebuttal{\caption{\label{fig:faceforensics_ffhq_pretrained_grid}{\bf{Qualitative results for pre-training on FFHQ before training on FaceForensics.}} We show qualitative results of our model pre-trained on FFHQ and then trained on FaceForensics, rendered from five different cameras (columns) for three identities (rows). We observe high view-consistency and high quality results.}}
    \vspace{-0.0em}
\end{figure*}

 \begin{figure*}[t]
    \centering
    \includegraphics[scale=0.2]{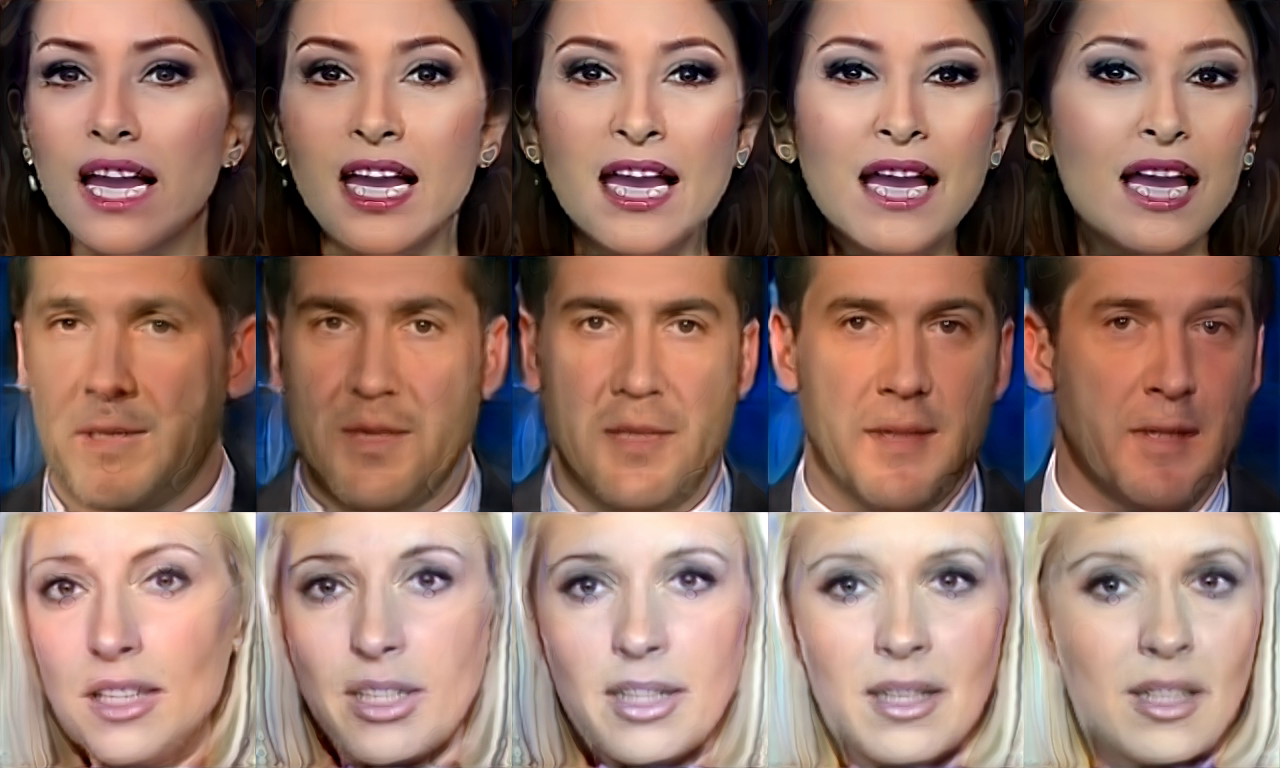}
    \vspace{-0.5em}
    \rebuttal{\caption{\label{fig:faceforensics_ffhq_simultan_grid}{\bf{FFHQ and FaceForensics simultaneous training results.}} We show qualitative results of our model simultaneously trained on FFHQ and FaceForensics, rendered from five different cameras (columns) for three identities (rows). We observe lower view-consistency compared to our model only pre-trained on FFHQ.}}
    \vspace{-0.0em}
\end{figure*}

\section{Training on Image and Video Datasets}
\label{sec:image_video_datasets}
\subsection{Training Setup}
In this section we discuss the use of an auxiliary 2D image dataset for training our 4D GANs, as described in Sec. \ref{sec:method_training} in the main paper. As motivated in the main paper, the lack of diversity of the provided videos could diminish the 3D accuracy of 4D GANs.
In fact, we observe that the model trained only on the FaceForensics dataset \citep{Rossler2019ICCV} exhibits narrow range of camera angles that generates view-consistent renderings.

As described in the main paper, we explored two options to leverage an image dataset to complement the video dataset: (i) pre-training on the image dataset, and (ii) simultaneously training both on image and video datasets. For both options, we follow the training setup of StyleNeRF \citep{Gu2022ICLR} on computing and backpropagating the image losses.

\rebuttal{In early experiments we also explored pre-training and simultaneous training for MEAD to better interpolate between the discrete poses in the data distribution. However, combining MEAD with FFHQ does not lead to any improvement in neither quality nor 3D behavior. We hypothesize that training on the image dataset mainly leads to improvements when the camera poses of the video dataset are already diverse but narrow as in FaceForensics. MEAD provides a wide but extremely sparse range of camera poses, which can not be interpolated by the image dataset distribution. We did not leverage image datasets for TaiChi as the viewpoint distribution is already diverse. For a fair comparison to previous works, we only used our vanilla approach for both qualitative and quantitative comparisons in our main paper.}
\rebuttal{\subsection{Qualitative Results}}
\rebuttal{The model pre-trained on the FFHQ image dataset and fine-tuned on the video dataset generates high quality, 3D-aware renderings, as can be seen on the supplementary videos. We generally observe sharper and higher quality renderings compared to our default model. Moreover, we notice that the range of allowed view angles increases, which is most obvious for the pitch (or elevation) angles, as described in Fig.~\ref{fig:pitch}. We note, however, that the colors seem less vivid compared to the pure-video model.}

\rebuttal{We further provide qualitative results for pre-training and simultaneous training in Fig.~\ref{fig:faceforensics_ffhq_pretrained_grid} and Fig.~\ref{fig:faceforensics_ffhq_simultan_grid}, respectively. For the case of simultaneous image and video training, the resulting model fails to output view-consistent videos. We hypothesize that training the generators to fit two different data distributions does not lead to consistent 4D models. However, only pre-training on FFHQ leads to view-consistent renderings while maintaining high quality.}
\rebuttal{\subsection{Quantitative Results}}
\rebuttal{In Table~\ref{tab:ffs_ffhq} we provide a quantitative comparison between i) pre-training on FFHQ before training on FaceForensics, ii) simultaneously training on FFHQ and FaceForensics, and iii) only training on FaceForensics as a vanilla approach. Generally, it is difficult to interpret FVD and FID when using two training datasets as these scores compare real and fake data distributions. Due to the usage of the FFHQ dataset the latent space of the model is manipulated with a dataset that is not used as part of the evaluation process. We still provide these two scores for completeness. For our pre-trained model, we observe that the FVD gap is significantly higher than the FID gap in comparison to our vanilla approach. We hypothesize that due to the model converging to a static dataset in the first training phase, namely FFHQ, the model has difficulties in learning motion in the second training phase with FaceForensics. However, we observe a significantly improved ID, which confirms the better multi-view consistency in our videos. Simultaneous training results in similar scores as our vanilla approach, however the multi-view consistency is significantly worse than using pre-training, which aligns with the lower ID score.}

\begin{table}[!t]
    \vspace{0.5em}
    \centering
    {\colrebuttal{}\begin{tabularx}{0.75\linewidth}{l|ccccc}
        \toprule
        \multirow{1}{*}{Method} & FVD ($\downarrow$) & FID ($\downarrow$) & ACD ($\downarrow$) & CPBD ($\uparrow$) & ID ($\uparrow$)\\
        \midrule
        \multirow{1}{*}{Pre-trained} & 127.2 & 17.2 & 0.819 & 0.2084 & 0.982 \\
        \multirow{1}{*}{Simultaneously} & 62.1 & 18.2 & 0.945 & 0.2077 & 0.893\\
        \multirow{1}{*}{Vanilla} & 68.7 & 13.7 & 0.965 & 0.196 & 0.861\\
        \bottomrule
    \end{tabularx}}
    \rebuttal{\caption{\label{tab:ffs_ffhq}{\bf{Quantitative Results on FaceForensics in combination with FFHQ.}} We report metrics for pre-training on FFHQ before training on FaceForensics, simultaneously training on FFHQ and FaceForensics, and only training on FaceForensics (vanilla).}}
    \vspace{-0.0cm}
\end{table}
\rebuttal{\section{\label{sec:ablations_appendix}Ablations}}
\rebuttal{We provide further details for the ablations conducted in Sec. \ref{sec:exp_ablations}.}

\rebuttal{\subsection{Motion Vector}}
\rebuttal{Generally, there are different ways to incorporate time into the motion generator. We use a simple multiplication due to our empirical results on TaiChi showing the best FVD score of 158.3 for this setup, as shown in Table \ref{tab:ablation_taichi_motion_vector}. As explained in Sec. \ref{sec:exp_ablations}, using positional encoding leads to a worse FVD with 175.4. Furthermore, we experimented with concatenation of time, which also results in a slightly worse FVD with 164.7. We leave more complex incorporation of time, like acyclic positional encoding \citep{Skorokhodov2022CVPR}, for future work.}

\begin{table}[!t]
     \centering
    \vspace{-0.0em}
    {\colrebuttal{}\begin{tabularx}{0.4\linewidth}{@{}X@{}c@{}}
        \toprule
        Method & FVD ($\downarrow$) \\
        \midrule
        w/ Positional Time Encoding  & 175.4 \\
        w/ Concatenation  & 164.7 \\
        w/ Multiplication  & 158.3  \\
        \midrule
    \end{tabularx}}
    \vspace{-0.5em}
    \rebuttal{\caption{\label{tab:ablation_taichi_motion_vector}{\bf{Ablation of time incorporation in motion generator on TaiChi Dataset.}} We choose simple multiplication for our default model due to its low FVD score in comparison to positional time encoding and and concatenation.}}
    \vspace{-0.5em}
\end{table}

\rebuttal{\subsection{Image Discriminator}}

\rebuttal{Theoretically, our video discriminator should suffice for the model to learn from videos. However, we observe that using a separate image discriminator significantly boosts the quality of our results, as shown in Table \ref{tab:ablation_image_discriminator}.
Removing the image discriminator leads to a FVD score of 234.3. Another option is to use a deterioration of our video discriminator as an image discriminator. For this, we repeat every image along the channel dimension to form a 6 channel tensor and set the time difference to zero, which we further concatenate to obtain a 7 channel tensor for our video discriminator. This leads to a FVD score of 190.9, which is substantially better than just removing the image discriminator, however worse than using a separate image discriminator (158.3). Consequently, we observe that using two discriminators with disentangled weights significantly boosts the quality.}

\begin{table}[!t]
     \centering
    \vspace{-0.0em}
    {\colrebuttal{}\begin{tabularx}{0.5\linewidth}{@{}X@{}c@{}}
        \toprule
        Method & FVD ($\downarrow$) \\
        \midrule
        w/o Image Discriminator  & 234.3 \\
        w/ Deterioration of Video Discriminator  & 190.9 \\
        w/ Separate Image Discriminator  & 158.3  \\
        \midrule
    \end{tabularx}}
    \vspace{-0.5em}
    \rebuttal{\caption{\label{tab:ablation_image_discriminator}{\bf{Ablation of image discriminator on TaiChi Dataset.}}}}
    \vspace{-0.5em}
\end{table}

\section{Related Works with Complete References}

In this section, we discuss, in more detail, relevant prior works that was omitted in our main paper due to space constraints.

\paragraph{Neural Implicit Representations}%
Neural Implicit Representation (NIR) \citep{Mescheder2019CVPR, Park2019CVPR, Chen2019CVPRb, Michalkiewicz2019ICCV} have demonstrated impressive results on various tasks due to their continuous, efficient, and differentiable representation. Among the most widely utilized coordinate-based representations are Neural Radiance Fields (NeRFs) \citep{Mildenhall2020ECCV} that combine an implicit neural network with volumetric rendering to enforce 3D consistency, while performing the novel view synthesis task.
Implicit-based representations have been proven beneficial for other tasks including but not limited to 3D reconstruction of objects \citep{Mescheder2019CVPR, Park2019CVPR, Chen2019CVPRb, Michalkiewicz2019ICCV, Oechsle2021ICCV, Gropp2020ICML, Saito2019ICCV, Atzmon2019NIPS,
Sitzmann2019NIPS} and scenes \citep{Jiang2020CVPR, Chibane2020CVPR,
Peng2020ECCV, Chabra2020ECCV, Sitzmann2020NIPS}, novel-view synthesis of static
\citep{Barron2021ICCV, Bergman2921NIPS, Gao2020ARXIV,
Jiang2020CVPR, Liu2020NIPS, Lindell2022CVPR,Srinivasan2021CVPR, Piala2021THREEDV,
Brualla2021CVPR, Oechsle2021ICCV, Sajjadi2022CVPR} and dynamic
\citep{Lombardi2019SIGGRAPH, Li2021CVPR, Pumarola2021CVPR, Park2021ICCV,
Xian2021CVPR, Yuan2021CVPR, Park2021SIGGRAPH, Tretschk2021ICCV,
Tewari2021ARXIV} environments, inverse graphics \citep{Niemeyer2020CVPR, Yariv2020NIPS, Lin2020NIPS} as well as for representing videos
\citep{Chen2021NIPS, Yu2022ICLR}.

\paragraph{3D-Aware Image Generations}%
Many recent models investigate how 3D representations can be incorporated in generative settings for improving the image quality
\citep{Park2017CVPR, Nguyen-Phuoc2018NIPS} and increasing the controllability
over various aspects of the image formation process \citep{Gadelha2017THREEDV,Chan2021CVPR, Henderson2019IJCV, Henzler2019ICCV, Henderson2020NIPS,
Liao2020CVPR, Lunz2020ARXIV,Nguyen-Phuoc2019ICCV, Nguyen-Phuoc2020ARXIV, Schwarz2020NIPS, Gu2022ICLR, DeVries2021ICCV, Hao2021ICCV,
Meng2021ICCV, Niemeyer2021THREEDV, Zhou2021ARXIV}. Towards this goal, \cite{Henzler2019ICCV} proposed a GAN-based architecture that combines
voxel-based representations with differentiable rendering. However, due to the
low voxel resolution, their generated images suffered from various artifacts. Concurrently, HoloGAN \citep{Nguyen-Phuoc2019ICCV} demonstrated that adding inductive biases about the 3D structure of the world allows control over the pose of the generated objects. \cite{Nguyen-Phuoc2020ARXIV} extend this by also considering
the compositional structure of the world into objects. While both
\cite{Nguyen-Phuoc2019ICCV, Nguyen-Phuoc2020ARXIV} demonstrated impressive
results, their performance was less consistent in higher resolutions.
Another line of work \cite{Liao2020CVPR} propose to combine a 3D generator
with a differentiable renderer and a 2D image generator to enable 3D
controllability. Despite their promising results in synthetically generated
environments, their model struggles to generalize in real-world scenarios.
More recent approaches propose generative models using MLP-based radiance fields. These methods use volume rendering to obtain images from the 3D fields to model single \citep{Schwarz2020NIPS,Chan2021CVPR} or multiple \citep{Niemeyer2021CVPR} objects. 
Similarly, \citet{Zhou2021ARXIV}, \citet{Chan2021CVPR}, and
\citet{DeVries2021ICCV} explored the idea of combining NeRF with GANs for designing
3D-aware image generators. \cite{xu2021generative} propose a transition from cumulative rendering to rendering with only the surface points. ShadeGAN \citep{pan2021shading} introduces a multi-lighting constraint to improve the 3D representation. More recently, StyleSDF \citep{Or2022CVPR} and StyleNeRF
\citep{Gu2022ICLR} proposed to combine an MLP-based volume renderer with a style-based generator \citep{Karras2020CVPR} to
produce high-resolution 3D-aware images. Moreover, VolumeGAN \citep{xu20223d} characterizes the underlying 3D structure with a feature volume. \citet{Deng2022CVPR} explored learning a generative radiance field on 2D manifolds and \citet{Chan2022CVPR} introduced a 3D-aware architecture that exploits both implicit and explicit representations. To improve view-consistency, EpiGRAF \citep{skorokhodov2022epigraf} proposes patch-based training to discard the 2D upsampling network, while VoxGRAF \citep{schwarz2022voxgraf} uses sparse voxel grids for efficient rendering without a superresolution module.

\paragraph{GAN-based Video Synthesis}%
Inspired by the success of GANs and adversarial training on
photorealistic image generation, researchers  shifted their attention
to various video synthesis tasks \citep{Saito2017ICCV, Tulyakov2018CVPR,
Acharya2018ARXIV, Clark2020ARXIV, Yushchenko2019ARXIV, Kahembwe2019ARXIV,
Aich20202CVPR, Saito2020IJCV, Gordon2020ARXIV, Holynski2021CVPR,Munoz2021WACV, Tian2021ICLR,
Fox2021ARXIV, Yan2021ARXIV, Hyun2021CVPR, Yu2022ICLR, Skorokhodov2022CVPR}. Several recent
works pose the video synthesis as an autoregressive video prediction
task and seek to generate discrete future frames conditioned on the previous using either
recurrent~\citep{Kalchbrenner2017ICML, Walker2021ARXIV} or
attention-based~\citep{Rakhimov2020ARXIV,
Weissenborn2020ICLR, Yan2021ARXIV} models. 
Other works on video generation
\citep{Saito2017ICCV, Tulyakov2018CVPR, Aich20202CVPR, Saito2020IJCV} tried to
disentangle the motion from the image
generation during the video synthesis process. While this paradigm has been
widely adopted, these approaches typically struggle to generate realistic
videos. To facilitate
generating high-quality frames, \citet{Tian2021ICLR} and \citet{Fox2021ARXIV} employed a
pre-trained image generator of \citet{Karras2020CVPR}.
Recently, LongVideoGAN \citep{brooks2022generating} has investigated synthesizing longer videos of more complex datasets.
Closely related to our method are the recent and concurrent work of DIGAN~\citep{Yu2022ICLR} and StlyeGAN-V~\citep{Skorokhodov2022CVPR} that generate videos at continuous time step, without conditioning on previous frames.
DIGAN~\citep{Yu2022ICLR} employs an NIR-based image generator
\citep{Skorokhodov2021CVPR} for learning continuous video synthesis and introduces two discriminators:
the first discriminates the realism of each frame
and the second operates on image pairs and seeks to determine the realism of
the motion. Similarly, StyleGAN-V~\citep{Skorokhodov2022CVPR} employs a style-based GAN
\citep{Karras2020CVPR} and a single discriminator that operates on sparsely sampled frames.

\paragraph{\rebuttal{Dynamic View Synthesis}}
\rebuttal{Given a video of a dynamic scene, dynamic view synthesis methods create novel views at arbitrary viewpoints and time steps. 
Neural Volumes \citep{lombardi2019neural} uses an encoder-decoder network with ray marching to render dynamic objects with a multi-view capture system.
STaR \citep{yuan2021star} trains on multi-view videos to reconstruct rigid object motion by complementing a static NeRF with a dynamic NeRF. Similarly, \citet{gao2021dynamic} introduce a dynamic network, which predicts scene flow to create a warped radiance field from monocular videos. Inspired by level set methods, HyperNeRF \citep{park2021hypernerf} models each frame as a nonplanar slice through a hyperdimensional NeRF.
Another line of works \citep{gafni2021dynamic, liu2021neural, noguchi2021neural, peng2021neural} model dynamic humans from videos based on radiance fields.
Other approaches \citep{tretschk2021non, pumarola2021d, park2021nerfies} encode the scene into a canonical space to then deform the canonical radiance field.
Recently, DyNeRF \citep{li2022neural} uses multi-view videos and proposes a novel hierarchical training scheme in combination with ray importance sampling.
In contrast to our method, the aforementioned works require videos with different viewpoints to train a single network per scene, hence are not able to generate novel scenes. Instead, we learn a generative model from unstructured single-view video collections.}

\section{Limitations \& Discussions (continued)}
As mentioned in the main paper, we inherit the features of existing 3D-aware GANs that typically work the best when there is a single target object at the center of the scenes, e.g., human faces. A possible way of modeling multi-object scenes is to add in compositionality to scene distribution modeling (as in \citet{Niemeyer2021CVPR} and \citet{Hudson2021ICMLR}). Another factor that limits the modeling of larger and more complex scene is  sampling of  virtual cameras. Training of GANs involves approximating the input data distribution with the generators. For the case of 3D GANs, image generation requires sampling virtual viewpoints from a plausible viewpoint distribution, which could be designed manually for single-object scenes with outside-inward looking cameras. However, for larger scenes, obtaining or designing the plausible viewpoint distribution becomes challenging, due to lack of correspondences across scenes and high structural variability. An interesting future direction would be to model the camera viewpoint distribution using neural networks. 


While our 4D GAN compares competitively in terms of training time until convergence, we note that the volume rendering process or our approach leaves huge memory footprints. In fact, training of our model consumes ``2.4 GB / batch size," while the 2D video method of StyleGAN-V \citep{Skorokhodov2022CVPR} only consumes ``0.8 GB / batch size."  This relatively high memory consumption prevents us from using larger models with the highest generation qualities and limits the resolution of output renderings. We leave exploration of more efficient implicit representations, such as tri-plane representations \citep{Chan2022CVPR}, as future work to improve the efficiency and quality 4D GANs.




\end{document}